\documentclass{article}
    
    \usepackage[preprint]{neurips_2026}
    \usepackage{amsmath,amssymb,amsfonts}
    \usepackage{graphicx}
    \usepackage{textcomp}
    \usepackage{xcolor}
    \usepackage[hyphens]{url}
    \usepackage{booktabs}
    \usepackage[colorlinks=true,urlcolor=blue,citecolor=blue,linkcolor=black]{hyperref}
    \usepackage{makecell}
    \usepackage{multirow}
    \usepackage{subcaption}
    \usepackage{float}
    \usepackage{nicefrac}
    \usepackage{tabularx}
    \usepackage{xspace}
    \usepackage{adjustbox}
    \usepackage{pifont}
    \usepackage[utf8]{inputenc}
    \usepackage[T1]{fontenc}
    \usepackage{microtype}
    \usepackage{enumitem}
    \usepackage{tikz}
    \usetikzlibrary{positioning,shapes.geometric,fit,calc,arrows.meta,backgrounds}
    \usepackage{type1cm}
    \usepackage{pgfplots}
    \pgfplotsset{compat=1.16}
    \usepackage[scaled=0.92]{helvet}
    \usepackage{cleveref}
    \usepackage[most]{tcolorbox}
    
    \tcbset{
      yamlbox/.style={
        enhanced,
        colback=gray!4,
        colframe=gray!45,
        arc=2mm,
        boxrule=0.5pt,
        left=1mm,
        right=1mm,
        top=1mm,
        bottom=1mm,
        fontupper=\scriptsize\ttfamily,
        before skip=0.5em,
        after skip=0.5em,
      }
    }
    
    \definecolor{okblue}{HTML}{0072B2}
    \definecolor{okorange}{HTML}{E69F00}
    \definecolor{okgreen}{HTML}{009E73}
    \definecolor{okred}{HTML}{D55E00}
    \definecolor{okpurple}{HTML}{CC79A7}
    \definecolor{okcyan}{HTML}{56B4E9}
    
    \newcommand{\sys}[0]{{\textsc{PerfReasoning}}\xspace}
    
    \newif\ifdraftcomments
    \draftcommentstrue

    \title{\sys: How Well Do LLMs Reason on Hardware Performance?}

    \author{
\textbf{Dan Zhao, Karthikeyan Sankaralingam, Christos Kozyrakis, Qijing Huang} \\ 
\\
NVIDIA
}

\begin{document}
    \maketitle
    
    \begin{abstract}
    Performance modeling is central to hardware design and software optimization,
    yet constructing these models requires structured reasoning about computation,
    data reuse, storage, and movement. We introduce \sys{}, a benchmark that
    evaluates LLMs both as direct performance reasoners and as generators of
    analytical performance-model code. Given workload, architecture, and mapping
    specifications, models compare mappings and predict off-chip traffic and buffer
    requirements. The strongest closed-source models exceed $90\%$ on
    reasoning-based Q\&A, and the best open-weight model reaches $82.4\%$. However, model
    construction is substantially harder: while GPT-5.6~Sol exceeds $80\%$ pass rate,
    all other model configurations average below $45\%$ and vary markedly
    across runs. Task-specific RL raises a 4B model's mapping-reasoning accuracy by
    $15.7$ points, whereas feedback-free multi-round self-revision prompting is not
    reliably effective. \sys{} exposes the gap between plausible architectural
    reasoning and reliable performance-model construction.
    We will publicly release the benchmark to support reproducible evaluation and track future progress.
    
    \end{abstract}
    
    \section{Introduction}
    \label{sec:introduction-4p}
    
    Performance models guide hardware design and software optimization before
    expensive simulation or implementation
    \cite{chen2018tvm,zheng2020ansor,parashar2019timeloop}. Given a workload, an architecture, and a mapping, an analytical model estimates quantities
    such as memory traffic, storage utilization, execution time, energy, and cost. These models
    are typically developed manually by domain experts
    \cite{gober2022champsim,binkert2011gem5,lowepower2020gem5,
    kim2016ramulator,jung2015dramsys,won2023astrasim,raj2025scalesim,
    parashar2019timeloop,wilton1996cacti,
    muralimanohar2007cacti6}. Learned predictors reduce manual effort, but require data and may not generalize outside the training distribution
    \cite{renda2020difftune,kumar2021data}.
    
    Frontier LLMs offer two complementary paths: reasoning directly about a
    design's performance, or generating analytical performance-model code from
    specifications. Both require the model to connect tiling, loop order, and tensor
    placement to data reuse and memory traffic, then relate those demands to compute, memory, and interconnect constraints. Errors can silently steer an
    optimization loop toward poor decisions, especially when measurements are
    unavailable. We therefore ask: \emph{Can LLMs reliably reason about performance
    and build analytical performance models?}
    
    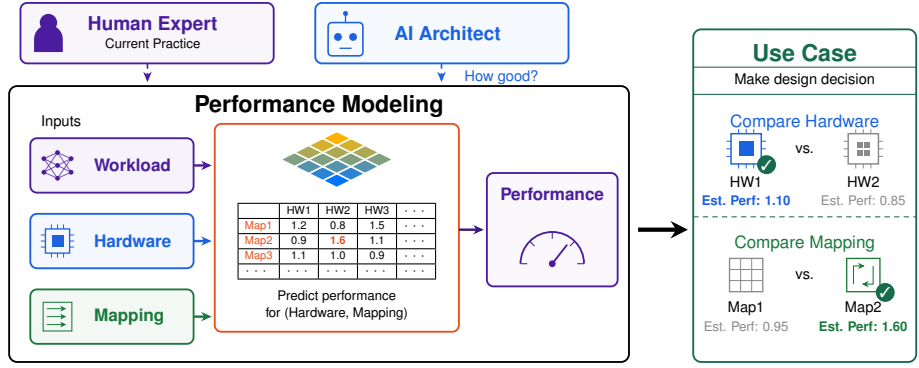
\begin{figure}[t]
      \centering
\providecommand{\ovtitle}{\sffamily\fontsize{9.5}{11}\selectfont\bfseries}
\providecommand{\ovhead}{\sffamily\fontsize{7.8}{9}\selectfont}
\providecommand{\ovlabel}{\sffamily\fontsize{7}{8.5}\selectfont}
\providecommand{\ovsmall}{\sffamily\fontsize{6}{7}\selectfont}
\providecommand{\ovtiny}{\sffamily\fontsize{5.4}{6.4}\selectfont}
\providecommand{\ovtable}{\sffamily\fontsize{4.5}{5.3}\selectfont}
\definecolor{ovpurple}{HTML}{4B1D9E}
\definecolor{ovblue}{HTML}{1B62E0}
\definecolor{ovgreen}{HTML}{1A7A3C}
\definecolor{ovorange}{HTML}{E8501E}
\definecolor{ovteal}{HTML}{176B4D}
\definecolor{ovgrey}{HTML}{8A8A8A}
\begin{tikzpicture}[
  font=\sffamily,
  panel/.style={rounded corners=3pt, draw=black, line width=0.7pt},
  card/.style={rounded corners=2.5pt, draw, line width=0.7pt},
  flow/.style={-{Stealth[length=4pt,width=3.5pt]}, line width=0.9pt},
  hint/.style={-{Stealth[length=4pt,width=3.2pt]}, line width=0.6pt,
               dash pattern=on 2pt off 1.6pt},
]

\draw[card, draw=ovpurple, fill=ovpurple!4] (0.15,4.40) rectangle (3.50,5.20);
\begin{scope}[shift={(0.52,4.80)}]                     
  \fill[ovpurple] (0,0.14) circle (0.115);
  \fill[ovpurple] (-0.19,-0.25) -- (-0.19,-0.08) arc (180:0:0.19) -- (0.19,-0.25) -- cycle;
\end{scope}
\node[font=\ovhead, text=ovpurple] at (1.90,4.91) {\bfseries Human Expert};
\node[font=\ovtiny] at (1.90,4.67) {Current Practice};

\draw[card, draw=ovblue, fill=ovblue!4] (4.05,4.40) rectangle (7.40,5.20);
\begin{scope}[shift={(4.45,4.66)}]                     
  \draw[ovblue, line width=0.5pt] (0,0.28) -- (0,0.37);
  \fill[ovblue] (0,0.40) circle (0.033);
  \draw[ovblue, line width=0.6pt, rounded corners=1.4pt] (-0.23,-0.16) rectangle (0.23,0.28);
  \fill[ovblue] (-0.10,0.09) circle (0.052);
  \fill[ovblue] (0.10,0.09) circle (0.052);
  \draw[ovblue, line width=0.5pt] (-0.08,-0.06) -- (0.08,-0.06);
\end{scope}
\node[font=\ovhead, text=ovblue] at (5.80,4.80) {\bfseries AI Architect};

\draw[panel] (0,0.45) rectangle (8.20,4.10);
\node[anchor=north, font=\ovtitle] at (4.10,4.10) {Performance Modeling};

\draw[hint, ovpurple] (1.83,4.40) -- (1.83,4.15);
\draw[hint, ovblue] (5.72,4.40) -- (5.72,4.15);
\node[anchor=west, font=\ovsmall, text=ovblue] at (5.90,4.22) {How good?};

\node[anchor=west, font=\ovsmall] at (0.30,3.62) {Inputs};
\foreach \cy/\col/\name in {3.05/ovpurple/Workload, 2.05/ovblue/Hardware,
                            1.05/ovgreen/Mapping} {
  \draw[card, draw=\col, fill=\col!5] (0.28,\cy-0.36) rectangle (2.44,\cy+0.36);
  \node[anchor=west, font=\ovlabel, text=\col] at (1.00,\cy) {\bfseries\name};
  \draw[flow, \col] (2.44,\cy) -- (2.68,\cy);
}
\begin{scope}[shift={(0.64,3.05)}]                     
  \foreach \a in {-0.09,0.09}                          
    \foreach \b in {-0.17,0,0.17}
      \draw[ovpurple, line width=0.22pt, opacity=0.75] (-0.18,\a) -- (0,\b);
  \foreach \b in {-0.17,0,0.17}                        
    \foreach \c in {-0.09,0.09}
      \draw[ovpurple, line width=0.22pt, opacity=0.75] (0,\b) -- (0.18,\c);
  \foreach \a in {-0.09,0.09} \fill[ovpurple] (-0.18,\a) circle (0.032);
  \foreach \b in {-0.17,0,0.17} \fill[ovpurple] (0,\b) circle (0.032);
  \foreach \c in {-0.09,0.09} \fill[ovpurple] (0.18,\c) circle (0.032);
\end{scope}
\begin{scope}[shift={(0.64,2.05)}]                     
  \draw[ovblue, line width=0.5pt] (-0.16,-0.16) rectangle (0.16,0.16);
  \fill[ovblue] (-0.075,-0.075) rectangle (0.075,0.075);
  \foreach \d in {-0.08,0,0.08} {
    \draw[ovblue, line width=0.32pt] (\d,0.16) -- (\d,0.22) (\d,-0.16) -- (\d,-0.22);
    \draw[ovblue, line width=0.32pt] (0.16,\d) -- (0.22,\d) (-0.16,\d) -- (-0.22,\d);
  }
\end{scope}
\begin{scope}[shift={(0.64,1.05)}]                     
  \draw[ovgreen, line width=0.4pt] (-0.19,-0.19) rectangle (0.19,0.19);
  \foreach \y in {-0.105,0,0.105}
    \draw[ovgreen, line width=0.32pt, -{Stealth[length=2.2pt,width=2pt]}]
      (-0.14,\y) -- (0.115,\y);
\end{scope}

\draw[card, draw=ovorange] (2.72,0.85) rectangle (5.95,3.60);
\begin{scope}[shift={(4.335,2.85)}, scale=0.30]
  \foreach \r in {0,1,2,3} {
    \foreach \c in {0,1,2,3} {
      \pgfmathsetmacro{\h}{(\r+\c)/6}
      \pgfmathsetmacro{\cr}{\h}
      \pgfmathsetmacro{\cg}{0.45+0.25*sin(90*\h)}
      \pgfmathsetmacro{\cb}{1-\h}
      \definecolor{cell}{rgb}{\cr,\cg,\cb}
      \pgfmathsetmacro{\cx}{(\c-\r)*0.62}
      \pgfmathsetmacro{\cy}{(\c+\r)*0.31}
      \fill[cell, draw=white, line width=0.6pt]
        (\cx-0.62,\cy) -- (\cx,\cy-0.31) -- (\cx+0.62,\cy) -- (\cx,\cy+0.31) -- cycle;
    }
  }
\end{scope}
\node[anchor=north, font=\ovtiny, align=center] at (4.335,2.68) {%
  \setlength{\tabcolsep}{2.4pt}\renewcommand{\arraystretch}{1.0}%
  {\ovtable\begin{tabular}{|l|c|c|c|c|}
    \hline
    & HW1 & HW2 & HW3 & $\cdots$\\\hline
    \textcolor{ovorange}{Map1} & 1.2 & 0.8 & 1.5 & $\cdots$\\\hline
    \textcolor{ovorange}{Map2} & 0.9 & \textcolor{ovorange}{\bfseries 1.6} & 1.1 & $\cdots$\\\hline
    \textcolor{ovorange}{Map3} & 1.1 & 1.0 & 0.9 & $\cdots$\\\hline
    $\cdots$ & $\cdots$ & $\cdots$ & $\cdots$ & $\cdots$\\\hline
  \end{tabular}}\\[3pt]
  Predict performance\\for (Hardware, Mapping)};

\draw[flow, ovpurple!80!black] (5.95,2.20) -- (6.28,2.20);

\draw[card, draw=ovpurple, fill=ovpurple!3] (6.32,1.45) rectangle (8.05,2.95);
\node[anchor=north, font=\ovlabel, text=ovpurple] at (7.185,2.88) {\bfseries Performance};
\begin{scope}[shift={(7.185,1.75)}]                    
  \draw[ovpurple, line width=0.7pt] (-0.42,0) arc (180:0:0.42);
  \foreach \a in {0,45,90,135,180}
    \draw[ovpurple, line width=0.4pt] (\a:0.42) -- (\a:0.33);
  \draw[ovpurple, line width=0.7pt] (0,0) -- (52:0.32);
  \fill[ovpurple] (0,0) circle (0.045);
\end{scope}

\draw[black, line width=1.6pt, -{Stealth[length=6.5pt,width=6.5pt]}]
  (8.32,2.20) -- (8.98,2.20);

\draw[panel, draw=ovteal] (9.05,0.45) rectangle (12.00,4.85);
\node[anchor=north, font=\ovtitle, text=ovteal] at (10.525,4.78) {Use Case};
\draw[ovteal, line width=0.4pt] (9.05,4.35) -- (12.00,4.35);
\node[font=\ovsmall] at (10.525,4.17) {Make design decision};
\draw[ovteal, line width=0.4pt] (9.05,3.99) -- (12.00,3.99);

\node[anchor=north, font=\ovlabel, text=ovblue] at (10.525,3.85) {Compare Hardware};
\begin{scope}[shift={(9.75,3.27)}]                     
  \draw[ovblue, line width=0.55pt] (-0.20,-0.20) rectangle (0.20,0.20);
  \fill[ovblue] (-0.095,-0.095) rectangle (0.095,0.095);
  \foreach \d in {-0.10,0,0.10} {
    \draw[ovblue, line width=0.32pt] (\d,0.20) -- (\d,0.26) (\d,-0.20) -- (\d,-0.26);
    \draw[ovblue, line width=0.32pt] (0.20,\d) -- (0.26,\d) (-0.20,\d) -- (-0.26,\d);
  }
  \fill[ovteal] (0.27,-0.21) circle (0.125);
  \node[white, font=\ovtiny] at (0.27,-0.21) {\ding{51}};
\end{scope}
\node[font=\ovsmall] at (10.525,3.27) {vs.};
\begin{scope}[shift={(11.30,3.27)}]                    
  \draw[ovgrey, line width=0.55pt] (-0.20,-0.20) rectangle (0.20,0.20);
  \foreach \x in {-0.088,0.012}                        
    \foreach \y in {-0.088,0.012}
      \fill[ovgrey] (\x,\y) rectangle (\x+0.076,\y+0.076);
  \foreach \d in {-0.10,0,0.10} {
    \draw[ovgrey, line width=0.32pt] (\d,0.20) -- (\d,0.26) (\d,-0.20) -- (\d,-0.26);
    \draw[ovgrey, line width=0.32pt] (0.20,\d) -- (0.26,\d) (-0.20,\d) -- (-0.26,\d);
  }
\end{scope}
\node[font=\ovsmall] at (9.75,2.83) {HW1};
\node[font=\ovsmall] at (11.30,2.83) {HW2};
\node[font=\ovtiny, text=ovblue] at (9.75,2.59) {\bfseries Est. Perf: 1.10};
\node[font=\ovtiny, text=ovgrey] at (11.30,2.59) {Est. Perf: 0.85};

\draw[ovteal, line width=0.4pt, dash pattern=on 2pt off 1.6pt] (9.05,2.37) -- (12.00,2.37);

\node[anchor=north, font=\ovlabel, text=ovgreen] at (10.525,2.25) {Compare Mapping};
\begin{scope}[shift={(9.75,1.60)}]                     
  \draw[ovgrey, line width=0.5pt] (-0.22,-0.22) rectangle (0.22,0.22);
  \draw[ovgrey, line width=0.35pt]
    (-0.0733,-0.22) -- (-0.0733,0.22) (0.0733,-0.22) -- (0.0733,0.22)
    (-0.22,-0.0733) -- (0.22,-0.0733) (-0.22,0.0733) -- (0.22,0.0733);
\end{scope}
\node[font=\ovsmall] at (10.525,1.60) {vs.};
\begin{scope}[shift={(11.30,1.60)}]                    
  \draw[ovgreen, line width=0.5pt] (-0.22,-0.22) rectangle (0.22,0.22);
  \draw[ovgreen, line width=0.4pt, -{Stealth[length=2.6pt,width=2.4pt]}]
    (-0.13,-0.12) -- (-0.13,0.12) -- (0.02,0.12);
  \draw[ovgreen, line width=0.4pt, -{Stealth[length=2.6pt,width=2.4pt]}]
    (0.13,0.12) -- (0.13,-0.12) -- (-0.02,-0.12);
  \fill[ovteal] (0.29,-0.23) circle (0.125);
  \node[white, font=\ovtiny] at (0.29,-0.23) {\ding{51}};
\end{scope}
\node[font=\ovsmall] at (9.75,1.16) {Map1};
\node[font=\ovsmall] at (11.30,1.16) {Map2};
\node[font=\ovtiny, text=ovgrey] at (9.75,0.92) {Est. Perf: 0.95};
\node[font=\ovtiny, text=ovgreen] at (11.30,0.92) {\bfseries Est. Perf: 1.60};

\end{tikzpicture}
      \caption{\sys{} evaluates LLMs as direct performance reasoners and as
      generators of analytical performance-model code.}
      \label{fig:overview-4p}
      \label{fig:map2perf-overview}
      \vspace{-0.2cm}
    \end{figure}

    We answer this question with \sys{}, which couples reasoning-based Q\&A with
    execution-based evaluation of generated models. This paired design separates
    performance understanding from the ability to turn that understanding
    into quantitatively reliable code. Across recent open-weight and closed-source
    models, we find strong comparative reasoning but a large and unstable gap in
    model construction. Moreover, code-induced mapping rankings closely match
    direct LLM comparisons on average, showing that generated programs can preserve
    useful ordinal signal even when their absolute predictions are wrong.
    
    \begin{minipage}{\linewidth}
    Prior AI benchmarks span systems code generation
    \cite{ouyang2025kernelbench,wen2025multikernelbench,
    wang2026kernelbenchx,lin2026solexecbench}, HLS and RTL
    \cite{abikaram2025hlseval,khan2026bench4hls,liu2023verilogeval,
    lu2023rtllm,liu2025openllmrtl,pinckney2025cvdp}, and chip and physical design
    \cite{yu2026chipbench,li2026pdagentbench}; QuArch~\cite{prakash2026quarch} evaluates
computer-architecture Q\&A, while ArchEval~\cite{wang2026archeval} evaluates broader tool-using
architecture agents.
Rosetta~\cite{sankaralingam2026rosetta} uses generated analytical models to audit paper claims; 
    Learning-based design-space exploration and emerging AI-architect flows depend
    on trusted evaluators to score candidate designs
    \cite{krishnan2023archgym,samajdar2021airchitect,architecture20,
    sankaralingam2026alphazero}. 
    \sys{} complements these efforts by isolating performance reasoning and analytical model construction in the benchmark with controlled labels.
    \end{minipage}
    
    \begin{minipage}{\linewidth}
    \paragraph{Contributions.}
    \begin{itemize}[leftmargin=*,nosep,topsep=2pt]
        \item We introduce \sys{}, evaluating LLMs as performance reasoners and
        generators of analytical performance model code.
        \item We evaluate open-weight and closed-source frontier models, revealing
        a large reasoning--construction gap and substantial run-to-run variance.
        \item We test task-specific RL and feedback-free multi-round self-revision:
        RL adds $15.7$ points, whereas self-revision is unreliable.
    \end{itemize}
    \end{minipage}
    \section{Benchmark}
    \label{sec:method-4p}
    \label{sec:methodology}
    
    \paragraph{Tasks.}
    Each Q\&A task in the benchmark presents two mappings using three prompt templates agreement with a
    correct statement, agreement with its negation, and a direct comparison asking
    which mapping is better. We report the minimum accuracy across formulations to
    control for agreement bias. Each pair isolates one mapping axis---tile size,
    loop order, or tensor keep/bypass---and compares off-chip traffic or buffer
    capacity, key inputs to analytical performance and energy models. 
    
    The construction task prompts each LLM to generate one Python analytical model.
    The program reads workload, architecture, and mapping YAML specifications and
    predicts total and per-tensor backing-store accesses and buffer capacity. We run
    the generated program on hidden cases and compare it with Timeloop-derived
    labels \cite{parashar2019timeloop}. A case passes when both access and capacity
    predictions agree within $10^{-6}$ relative error; we also report Q-error and
    pairwise preference accuracy. The latter asks only whether the generated model
    ranks two mappings in the same order as ground truth.
    ranks two mappings in the same order as the reference model.
    
    \paragraph{Coverage and Setup.}
    The benchmark contains 216 valid configurations: 108 paired mappings spanning
    matrix multiplication, batched matrix multiplication, and convolution. The
    fixed two-level architecture is MainMemory $\rightarrow$ Buffer $\rightarrow$
    MACC; mappings vary temporal tiling, loop permutation, and tensor retention.
    Labels span more than seven orders of magnitude in memory traffic and include
    both all-kept and bypass configurations. We evaluate 16 model families released
    or available from April-July 2026, including Claude, DeepSeek, Gemini, GLM,
    GPT, GPT-OSS, and MiniMax variants, at each provider-supported reasoning effort.
    We sample repeated generations when available and execute each program in an
    isolated environment. Appendix~\ref{app:details-4p} gives full task and setup details.
    
    \section{Results}
    \label{sec:results-4p}
    \label{sec:results}
    
    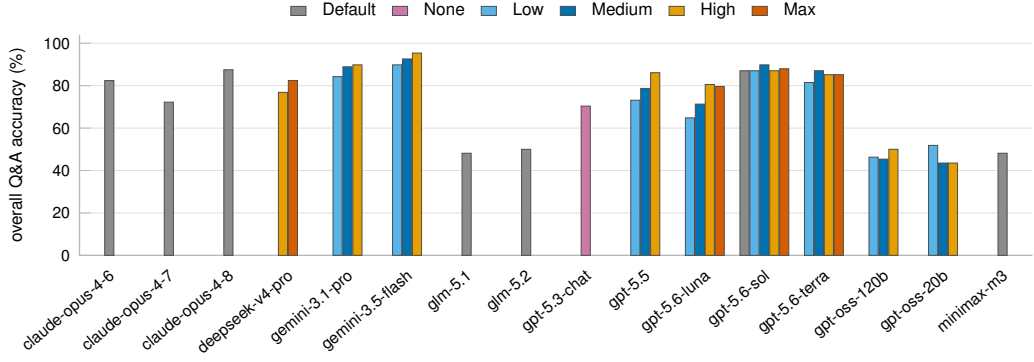
\begin{figure}[t]
      \centering
\begin{tikzpicture}[font=\sffamily]
\begin{axis}[
  x=0.1313cm, height=4.5cm,
  xmin=-3.00, xmax=93.00, ymin=0, ymax=104,
  ytick={0,20,40,60,80,100},
  ylabel={overall Q\&A accuracy (\%)},
  ylabel style={font=\sffamily\fontsize{7}{8}\selectfont},
  xtick={0.000,6.000,12.000,18.000,24.000,30.000,36.000,42.000,48.000,54.000,60.000,66.000,72.000,78.000,84.000,90.000},
  xticklabels={{claude-opus-4-6},{claude-opus-4-7},{claude-opus-4-8},{deepseek-v4-pro},{gemini-3.1-pro},{gemini-3.5-flash},{glm-5.1},{glm-5.2},{gpt-5.3-chat},{gpt-5.5},{gpt-5.6-luna},{gpt-5.6-sol},{gpt-5.6-terra},{gpt-oss-120b},{gpt-oss-20b},{minimax-m3}},
  x tick label style={font=\sffamily\fontsize{6.5}{8}\selectfont, rotate=40, anchor=north east},
  y tick label style={font=\sffamily\fontsize{6.5}{8}\selectfont},
  /pgf/number format/assume math mode=true,
  ymajorgrids, grid style={gray!25, line width=0.3pt},
  axis line style={gray!55, line width=0.5pt}, tick style={gray!55},
  axis x line*=bottom, axis y line*=left,
  legend style={at={(0.5,1.02)}, anchor=south, legend columns=6, draw=none,
                fill=none, font=\sffamily\fontsize{7}{8}\selectfont, column sep=0.9ex},
  legend image code/.code={\draw[#1] (0cm,-0.06cm) rectangle (0.20cm,0.06cm);},
]
\addplot[ybar, bar shift=0pt, bar width=3.44pt, fill=black!45, draw=black!70, line width=0.25pt] coordinates {(0.000,82.407) (6.000,72.222) (12.000,87.500) (36.000,48.148) (42.000,50.000) (64.000,87.037) (90.000,48.148)};
\addlegendentry{Default}
\addplot[ybar, bar shift=0pt, bar width=3.44pt, fill=okpurple, draw=black!70, line width=0.25pt] coordinates {(48.000,70.370)};
\addlegendentry{None}
\addplot[ybar, bar shift=0pt, bar width=3.44pt, fill=okcyan, draw=black!70, line width=0.25pt] coordinates {(23.000,84.259) (29.000,89.815) (53.000,73.148) (58.500,64.815) (65.000,87.037) (70.500,81.481) (77.000,46.296) (83.000,51.852)};
\addlegendentry{Low}
\addplot[ybar, bar shift=0pt, bar width=3.44pt, fill=okblue, draw=black!70, line width=0.25pt] coordinates {(24.000,88.889) (30.000,92.593) (54.000,78.704) (59.500,71.296) (66.000,89.815) (71.500,87.037) (78.000,45.370) (84.000,43.519)};
\addlegendentry{Medium}
\addplot[ybar, bar shift=0pt, bar width=3.44pt, fill=okorange, draw=black!70, line width=0.25pt] coordinates {(17.500,76.852) (25.000,89.815) (31.000,95.370) (55.000,86.111) (60.500,80.556) (67.000,87.037) (72.500,85.185) (79.000,50.000) (85.000,43.519)};
\addlegendentry{High}
\addplot[ybar, bar shift=0pt, bar width=3.44pt, fill=okred, draw=black!70, line width=0.25pt] coordinates {(18.500,82.407) (61.500,79.630) (68.000,87.963) (73.500,85.185)};
\addlegendentry{Max}
\end{axis}
\end{tikzpicture}
      \caption{Q\&A accuracy by model and reasoning effort.
      Strong closed models exceed $90\%$; the best open-weight result is $82.4\%$.}
      \label{fig:qa-4p}
      \label{fig:qa_overall}
    \end{figure}
    
    \paragraph{Frontier models reason well about mappings.}
    Figure~\ref{fig:qa-4p} shows that the strongest closed-source configurations
    reach $92.6$--$95.4\%$ Q\&A accuracy, while DeepSeek-V4-Pro, the strongest
    open-weight model, reaches $82.4\%$. Loop order is the most discriminative
    mapping axis: strong models distinguish reuse-sensitive permutations, whereas
    weaker models often treat them as equivalent. The appendix reports per-axis,
    workload, historical, and fine-tuning breakdowns.
    
    \begin{figure}[t]
      \centering
\begin{tikzpicture}[font=\sffamily]
\begin{axis}[
  x=0.1313cm, height=4.50cm,
  xmin=-3.00, xmax=93.00,
  ymin=0, ymax=104,
  /pgf/number format/assume math mode=true,
  ytick={0,20,40,60,80,100},
  ylabel={exact pass rate (\%)},
  ylabel style={font=\sffamily\fontsize{7}{8}\selectfont},
  xtick={0.000,6.000,12.000,18.000,24.000,30.000,36.000,42.000,48.000,54.000,60.000,66.000,72.000,78.000,84.000,90.000},
  xticklabels={{claude-opus-4-6},{claude-opus-4-7},{claude-opus-4-8},{deepseek-v4-pro},{gemini-3.1-pro},{gemini-3.5-flash},{glm-5.1},{glm-5.2},{gpt-5.3-chat},{gpt-5.5},{gpt-5.6-luna},{gpt-5.6-sol},{gpt-5.6-terra},{gpt-oss-120b},{gpt-oss-20b},{minimax-m3}},
  x tick label style={font=\sffamily\fontsize{6.5}{8}\selectfont, rotate=40, anchor=north east},
  y tick label style={font=\sffamily\fontsize{6.5}{8}\selectfont},
  ymajorgrids, grid style={gray!25, line width=0.3pt},
  axis line style={gray!55, line width=0.5pt}, tick style={gray!55},
  axis x line*=bottom, axis y line*=left,
  legend style={at={(0.5,1.02)}, anchor=south, legend columns=6, draw=none,
                fill=none, font=\sffamily\fontsize{7}{8}\selectfont, column sep=0.9ex},
  legend image code/.code={\draw[#1] (0cm,-0.06cm) rectangle (0.20cm,0.06cm);},
]
\addplot[ybar, bar shift=0pt, bar width=3.44pt, fill=okpurple, draw=black!70, line width=0.25pt, error bars/.cd, y dir=both, y explicit, error bar style={black!65, line width=0.4pt}, error mark options={mark size=0.8pt, black!65}] coordinates {(36.000,3.2) +- (0,0) (42.000,0) +- (0,0) (48.000,2.2) +- (0,1.1) (64.000,4.8) +- (0,4.6) (90.000,0) +- (0,0)};
\addlegendentry{None}
\addplot[ybar, bar shift=0pt, bar width=3.44pt, fill=okcyan, draw=black!70, line width=0.25pt, error bars/.cd, y dir=both, y explicit, error bar style={black!65, line width=0.4pt}, error mark options={mark size=0.8pt, black!65}] coordinates {(-1.000,0.2) +- (0,0.2) (5.000,4.6) +- (0,3) (11.500,8.3) +- (0,5.1) (23.000,1.2) +- (0,0.8) (29.000,6.2) +- (0,1.3) (53.000,0) +- (0,0) (58.500,0.3) +- (0,0.3) (65.000,55.2) +- (0,27.8) (70.500,1.2) +- (0,1.2) (77.000,2.7) +- (0,0.5) (83.000,2.1) +- (0,2.1)};
\addlegendentry{Low}
\addplot[ybar, bar shift=0pt, bar width=3.44pt, fill=okblue, draw=black!70, line width=0.25pt, error bars/.cd, y dir=both, y explicit, error bar style={black!65, line width=0.4pt}, error mark options={mark size=0.8pt, black!65}] coordinates {(0.000,0.3) +- (0,0.3) (6.000,17.7) +- (0,14) (24.000,4.8) +- (0,4.8) (30.000,38.4) +- (0,7.8) (54.000,0) +- (0,0) (59.500,0) +- (0,0) (66.000,84.7) +- (0,3.7) (71.500,0) +- (0,0) (78.000,3.4) +- (0,2) (84.000,0.2) +- (0,0.2)};
\addlegendentry{Medium}
\addplot[ybar, bar shift=0pt, bar width=3.44pt, fill=okorange, draw=black!70, line width=0.25pt, error bars/.cd, y dir=both, y explicit, error bar style={black!65, line width=0.4pt}, error mark options={mark size=0.8pt, black!65}] coordinates {(1.000,11.9) +- (0,11.9) (7.000,25.8) +- (0,13) (12.500,35.6) +- (0,20.1) (17.500,2.2) +- (0,0.7) (25.000,36.7) +- (0,13.7) (31.000,40) +- (0,14.3) (55.000,44.2) +- (0,25.5) (60.500,0) +- (0,0) (67.000,84.7) +- (0,3.7) (72.500,29.5) +- (0,29.5) (79.000,4.6) +- (0,2.4) (85.000,2.6) +- (0,2)};
\addlegendentry{High}
\addplot[ybar, bar shift=0pt, bar width=3.44pt, fill=okred, draw=black!70, line width=0.25pt, error bars/.cd, y dir=both, y explicit, error bar style={black!65, line width=0.4pt}, error mark options={mark size=0.8pt, black!65}] coordinates {(18.500,2.2) +- (0,1.1) (61.500,0) +- (0,0) (68.000,88.4) +- (0,0) (73.500,18.7) +- (0,13.7)};
\addlegendentry{Max}
\end{axis}
\end{tikzpicture}
      \caption{Single-shot pass rate for generated analytical models. Bars show
      means and error bars show variation across repeated generations.}
      \label{fig:exact-4p}
    \end{figure}
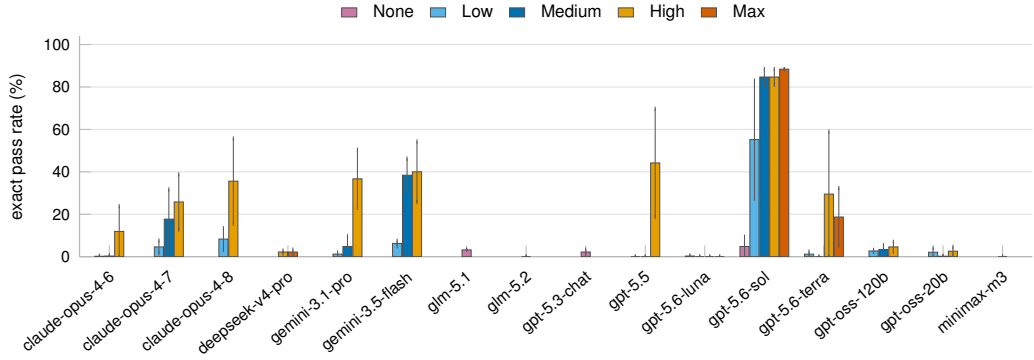
    
    \paragraph{Constructing a reliable model is much harder.}
    Figure~\ref{fig:exact-4p} reveals a sharp separation from Q\&A. GPT-5.6~Sol is
    the only family with consistently high performance, reaching $84.7$--$88.4\%$
    at medium through max effort. Every other model/effort configuration averages
    below $45\%$, and their aggregate average is below $15\%$. Buffer capacity is
    structurally easier than memory-access accounting; incorrect access predictions
    can miss by many orders of magnitude. Repeated samples also vary substantially,
    so a plausible program from one run does not imply a reliable evaluator.
    
    \paragraph{Code preserves rankings, not necessarily values.}
    \begin{figure}[t]
      \centering
\begin{tikzpicture}[font=\sffamily]
\begin{axis}[
  x=0.1313cm, height=4.50cm,
  xmin=-3.00, xmax=93.00,
  ymin=0, ymax=112,
  /pgf/number format/assume math mode=true,
  ytick={0,20,40,60,80,100},
  ylabel={pairwise preference accuracy (\%)},
  ylabel style={font=\sffamily\fontsize{7}{8}\selectfont},
  xtick={0.000,6.000,12.000,18.000,24.000,30.000,36.000,42.000,48.000,54.000,60.000,66.000,72.000,78.000,84.000,90.000},
  xticklabels={{claude-opus-4-6},{claude-opus-4-7},{claude-opus-4-8},{deepseek-v4-pro},{gemini-3.1-pro},{gemini-3.5-flash},{glm-5.1},{glm-5.2},{gpt-5.3-chat},{gpt-5.5},{gpt-5.6-luna},{gpt-5.6-sol},{gpt-5.6-terra},{gpt-oss-120b},{gpt-oss-20b},{minimax-m3}},
  x tick label style={font=\sffamily\fontsize{6.5}{8}\selectfont, rotate=40, anchor=north east},
  y tick label style={font=\sffamily\fontsize{6.5}{8}\selectfont},
  ymajorgrids, grid style={gray!25, line width=0.3pt},
  axis line style={gray!55, line width=0.5pt}, tick style={gray!55},
  axis x line*=bottom, axis y line*=left,
  legend style={at={(0.5,1.02)}, anchor=south, legend columns=6, draw=none,
                fill=none, font=\sffamily\fontsize{7}{8}\selectfont, column sep=0.9ex},
  legend image code/.code={\draw[#1] (0cm,-0.06cm) rectangle (0.20cm,0.06cm);},
]
\addplot[ybar, bar shift=0pt, bar width=3.44pt, fill=okpurple, draw=black!70, line width=0.25pt, error bars/.cd, y dir=both, y explicit, error bar style={black!65, line width=0.4pt}, error mark options={mark size=0.8pt, black!65}] coordinates {(36.000,59.6) +- (0,4.60901) (42.000,67) +- (0,7.29728) (48.000,65.7) +- (0,2.13833) (64.000,65.1) +- (0,2.22565) (90.000,53.7) +- (0,0)};
\addlegendentry{None}
\addplot[ybar, bar shift=0pt, bar width=3.44pt, fill=okcyan, draw=black!70, line width=0.25pt, error bars/.cd, y dir=both, y explicit, error bar style={black!65, line width=0.4pt}, error mark options={mark size=0.8pt, black!65}] coordinates {(-1.000,69.4) +- (0,6.67695) (5.000,71.9) +- (0,5.27408) (11.500,67.9) +- (0,3.26636) (23.000,70.4) +- (0,2.597) (29.000,77.2) +- (0,1.75139) (53.000,87.8) +- (0,6.24075) (58.500,62.9) +- (0,6.49628) (65.000,100) +- (0,0) (70.500,80.6) +- (0,1.85185) (77.000,57.7) +- (0,2.23845) (83.000,54.6) +- (0,3.29843)};
\addlegendentry{Low}
\addplot[ybar, bar shift=0pt, bar width=3.44pt, fill=okblue, draw=black!70, line width=0.25pt, error bars/.cd, y dir=both, y explicit, error bar style={black!65, line width=0.4pt}, error mark options={mark size=0.8pt, black!65}] coordinates {(0.000,62.7) +- (0,1.54321) (6.000,82.7) +- (0,0.308642) (24.000,87.3) +- (0,4.93827) (30.000,87) +- (0,4.17523) (54.000,94.1) +- (0,5.8642) (59.500,70.5) +- (0,1.08025) (66.000,100) +- (0,0) (71.500,94.1) +- (0,5.8642) (78.000,63.3) +- (0,1.23457) (84.000,48.6) +- (0,5.15534)};
\addlegendentry{Medium}
\addplot[ybar, bar shift=0pt, bar width=3.44pt, fill=okorange, draw=black!70, line width=0.25pt, error bars/.cd, y dir=both, y explicit, error bar style={black!65, line width=0.4pt}, error mark options={mark size=0.8pt, black!65}] coordinates {(1.000,82.9) +- (0,0.462963) (7.000,77.8) +- (0,5.0996) (12.500,72.5) +- (0,13.0106) (17.500,66) +- (0,2.89532) (25.000,82.9) +- (0,6.08466) (31.000,77.3) +- (0,6.30605) (55.000,95.6) +- (0,4.39815) (60.500,100) +- (0,0) (67.000,96.9) +- (0,3.08642) (72.500,100) +- (0,0) (79.000,59.3) +- (0,2.97644) (85.000,50.9) +- (0,3.8132)};
\addlegendentry{High}
\addplot[ybar, bar shift=0pt, bar width=3.44pt, fill=okred, draw=black!70, line width=0.25pt, error bars/.cd, y dir=both, y explicit, error bar style={black!65, line width=0.4pt}, error mark options={mark size=0.8pt, black!65}] coordinates {(18.500,63.3) +- (0,1.23457) (61.500,81.9) +- (0,0) (68.000,100) +- (0,0) (73.500,75.6) +- (0,7.25801)};
\addlegendentry{Max}
\end{axis}
\end{tikzpicture}
      \caption{Mapping-ranking accuracy induced by generated analytical models.
      Programs often preserve the correct ordering despite inaccurate values.}
      \label{fig:pairwise-main-4p}
    \end{figure}
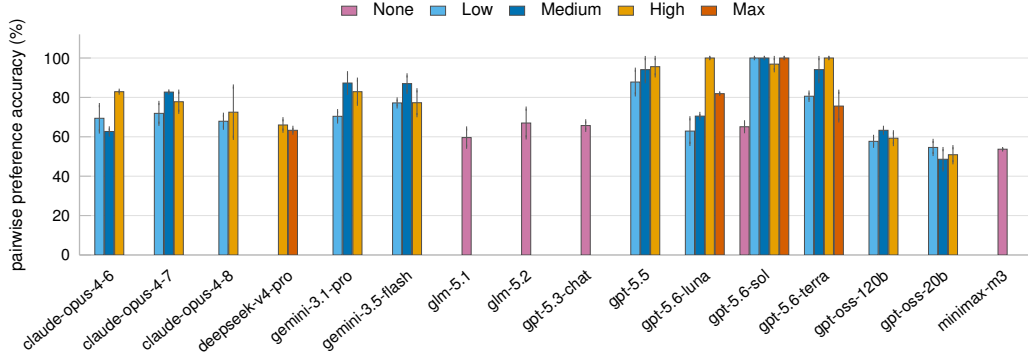
    
    Direct Q\&A and generated-model ranking evaluate the same ordinal question:
    which mapping moves less data? Across 31 aligned model/effort configurations,
    direct comparison and code-induced ranking average $78.3\%$ and $78.5\%$,
    respectively; code is higher in 15 configurations and lower in 16. Thus,
    translation into code preserves the comparative signal on average but does not
    systematically improve it. This is useful for selecting between candidates,
    but it cannot establish trustworthy absolute traffic or buffer estimates.
    
    \begin{figure}[t]
      \centering
      \begin{minipage}[c]{0.26\linewidth}
        \centering
        \begin{tikzpicture}[font=\sffamily]
          \begin{axis}[
            scale only axis=true,
            width=0.65\linewidth,
            height=2.30cm,
            ymin=0, ymax=80,
            xmin=0, xmax=1,
            ytick={0,20,40,60,80},
            xtick={0.35,0.65},
            xticklabels={Base,+RL},
            tick label style={font=\sffamily\fontsize{6.5}{8}\selectfont},
            title={Qwen3-4B},
            title style={font=\sffamily\fontsize{7.5}{9}\selectfont,yshift=-2pt},
            ylabel={Q\&A accuracy (\%)},
            ylabel style={font=\sffamily\fontsize{7}{8}\selectfont},
            ymajorgrids,
            grid style={gray!25,line width=0.3pt},
            axis line style={gray!55,line width=0.5pt},
            tick style={gray!55},
            axis x line*=bottom,
            axis y line*=left,
            nodes near coords,
            nodes near coords style={font=\sffamily\fontsize{6.5}{8}\selectfont},
          ]
          \addplot[ybar,bar width=10pt,fill=okgreen,draw=black!70]
            coordinates {(0.35,54.3)};
          \addplot[ybar,bar width=10pt,fill=okorange,draw=black!70]
            coordinates {(0.65,70.0)};
          \end{axis}
        \end{tikzpicture}
        \caption{Verifier-guided RL improves Qwen3-4B mapping-reasoning accuracy
        from $54.3\%$ to $70.0\%$.}
        \label{fig:rl-main-4p}
      \end{minipage}
      \hfill
      \begin{minipage}[c]{0.69\linewidth}
        \centering
        \resizebox{\linewidth}{!}{
\begin{tikzpicture}[font=\sffamily]
\begin{axis}[
  name=panel0,
  scale only axis=true,
  width=0.38\linewidth, height=2.30cm,
  xmin=-0.25, xmax=3.25, ymin=-4, ymax=100,
  xtick={0,1,2,3}, ytick={0,20,40,60,80,100},
  /pgf/number format/assume math mode=true,
  xlabel={self-revision round $k$},
  xlabel style={font=\sffamily\fontsize{7.5}{9}\selectfont},
  ylabel style={font=\sffamily\fontsize{7.5}{9}\selectfont},
  tick label style={font=\sffamily\fontsize{7}{8.5}\selectfont},
  title={(a) \textsf{gpt-5.6-sol} --- strongest closed},
  title style={font=\sffamily\fontsize{7.5}{9}\selectfont, yshift=-2pt},
  ymajorgrids, grid style={gray!25, line width=0.3pt},
  axis line style={gray!55, line width=0.5pt}, tick style={gray!55},
  axis x line*=bottom, axis y line*=left,
  ylabel={both-exact pass rate (\%)},
  legend style={at={(1.10,1.20)}, anchor=south, legend columns=4,
                draw=none, fill=none, font=\sffamily\fontsize{7.5}{9}\selectfont, column sep=0.9ex},
  legend image code/.code={\draw[#1, solid] (0cm,-0.06cm) rectangle (0.20cm,0.06cm);},
]
\addplot[okcyan, line width=0.9pt, mark=*, mark size=1.5pt, mark options={solid, fill=okcyan, draw=okcyan}] coordinates {(0,88.4259) (1,88.4259) (2,0.0000) (3,0.0000)};
\addlegendentry{Low}
\addplot[okblue, line width=0.9pt, mark=*, mark size=1.5pt, mark options={solid, fill=okblue, draw=okblue}, dash pattern=on 4pt off 1.6pt] coordinates {(0,88.4259) (1,88.4259) (2,88.4259) (3,88.4259)};
\addlegendentry{Medium}
\addplot[okorange, line width=0.9pt, mark=*, mark size=1.5pt, mark options={solid, fill=okorange, draw=okorange}, dash pattern=on 2.2pt off 1.8pt] coordinates {(0,88.4259) (1,88.4259) (2,88.4259) (3,88.4259)};
\addlegendentry{High}
\addplot[okred, line width=0.9pt, mark=*, mark size=1.5pt, mark options={solid, fill=okred, draw=okred}, dash pattern=on 1.2pt off 1.6pt] coordinates {(0,0.0000) (1,0.0000) (2,0.0000) (3,88.4259)};
\addlegendentry{Max}
\end{axis}
\begin{axis}[
  name=panel1,
  at={(panel0.east)}, anchor=west, xshift=1.0cm,
  scale only axis=true,
  width=0.38\linewidth, height=2.30cm,
  xmin=-0.25, xmax=3.25, ymin=-4, ymax=100,
  xtick={0,1,2,3}, ytick={0,20,40,60,80,100},
  /pgf/number format/assume math mode=true,
  xlabel={self-revision round $k$},
  xlabel style={font=\sffamily\fontsize{7.5}{9}\selectfont},
  ylabel style={font=\sffamily\fontsize{7.5}{9}\selectfont},
  tick label style={font=\sffamily\fontsize{7}{8.5}\selectfont},
  title={(b) \textsf{gpt-oss-20b} --- strongest open-weight},
  title style={font=\sffamily\fontsize{7.5}{9}\selectfont, yshift=-2pt},
  ymajorgrids, grid style={gray!25, line width=0.3pt},
  axis line style={gray!55, line width=0.5pt}, tick style={gray!55},
  axis x line*=bottom, axis y line*=left,
  yticklabels={},
]
\addplot[okcyan, line width=0.9pt, mark=*, mark size=1.5pt, mark options={solid, fill=okcyan, draw=okcyan}, forget plot] coordinates {(0,0.0000) (1,0.0000) (2,0.0000) (3,0.0000)};
\addplot[okblue, line width=0.9pt, mark=*, mark size=1.5pt, mark options={solid, fill=okblue, draw=okblue}, dash pattern=on 4pt off 1.6pt, forget plot] coordinates {(0,0.0000) (1,0.0000) (2,15.7407) (3,0.0000)};
\addplot[okorange, line width=0.9pt, mark=*, mark size=1.5pt, mark options={solid, fill=okorange, draw=okorange}, dash pattern=on 2.2pt off 1.8pt, forget plot] coordinates {(0,12.9630) (1,3.2407) (2,3.2407) (3,3.2407)};
\end{axis}
\end{tikzpicture}}
        \caption{Feedback-free multi-round self-revision is inconsistent for the
        strongest closed and open-weight models.}
        \label{fig:self-revision-main-4p}
      \end{minipage}
    \end{figure}
    
    \begin{minipage}{\linewidth}
    \paragraph{Task-specific RL improves mapping reasoning.}
    Figure~\ref{fig:rl-main-4p} compares the same Qwen3-4B backbone before and
    after verifier-guided RL on binary mapping questions. Accuracy rises from
    $54.3\%$ to $70.0\%$, showing that additional domain training can improve the
    reasoning prerequisite, although its effect on full model construction remains
    open.
    \end{minipage}
    
    \paragraph{Multi-round self-revision is not reliably effective.}
    We prompt each model to inspect and revise its performance-model code for up to
    three rounds without execution or pass/fail feedback.
    Figure~\ref{fig:self-revision-main-4p} shows representative trajectories;
    across all 41 configurations, 18 never change, 10 finish worse, and only 7
    finish better. Mean pass rate rises from $14.0\%$ to just $16.7\%$, showing
    that models cannot reliably recognize and retain improvements. Full results
    are reported in Appendix~\ref{app:iterative}.
    
    \paragraph{Implications and limitations.}
    The best frontier models exceed $90\%$ accuracy when reasoning about tiling,
    data reuse, and placement effects, making them promising for agentic design
    loops. However, variability across runs requires validation before
    their outputs can serve as reliable evaluators.
    
    Our benchmark evaluates frontier models out of the box through a
    natural-language interface and does not establish how complex a performance
    model frontier models can construct. 
    Formal verification tools, as used by Aristotle
    \cite{achim2025aristotle}, may enable more complex and reliable
    performance-model construction; we leave this to future work.
    Finally, Timeloop-derived labels measure analytical agreement rather than
    silicon accuracy; broader architectures and hardware-validated labels remain
    future work.
    

    \section{Conclusion}
    \label{sec:conclusion-4p}
    
    We introduced \sys{}, a benchmark of frontier LLMs as hardware-performance
    reasoners and builders of analytical model code. The strongest models reason
    accurately about mapping effects, yet only one evaluated configuration builds
    reliable models, and outcomes vary substantially across runs. Task-specific RL
    improves small-model reasoning, whereas feedback-free self-revision is
    unreliable. Progress toward AI-assisted chip design requires stronger reasoning
    and stable, accurate, and verifiable model generation. By separating reasoning
    from model construction, \sys{} provides a concrete benchmark for tracking
    that progress.
    
    \clearpage
    \bibliographystyle{plainnat}
    \bibliography{refs}
    
    \clearpage
    \appendix
    
    \section{Benchmark and Evaluation Details}
    \label{app:details-4p}
    
    Each generated program receives exactly three files: \texttt{prob.yaml}, which
    defines the tensor problem; \texttt{arch.yaml}, which defines the two-level
    hardware hierarchy; and \texttt{map.yaml}, which specifies temporal tiling,
    loop order, and keep/bypass decisions. The hidden \texttt{access.yaml} contains
    the reference result. The 216 cases are balanced across matrix multiplication,
    batched matrix multiplication, and convolution. Timeloop ground-truth access
    counts range from $98{,}304$ to $4{,}389{,}456{,}576{,}512$, and required buffer
    capacity ranges from 0 to $270{,}532{,}608$ elements.
    
    For Q\&A, the benchmark contains 108 mapping pairs in a balanced $3\times3$
    design over workload family and changed mapping axis. Each pair appears in
    three formulations. For model construction, each output must parse unseen YAML
    cases rather than hard-code benchmark answers. We record compile and runtime
    failures, exact access and capacity agreement, per-output Q-error, and paired
    ranking accuracy. Results are aggregated by model and reasoning-effort setting;
    error bars denote variation across repeated generated programs.
    
    \section{Benchmark Case Example and Components}
    \label{app:benchmark}
    
\begin{figure}[t]
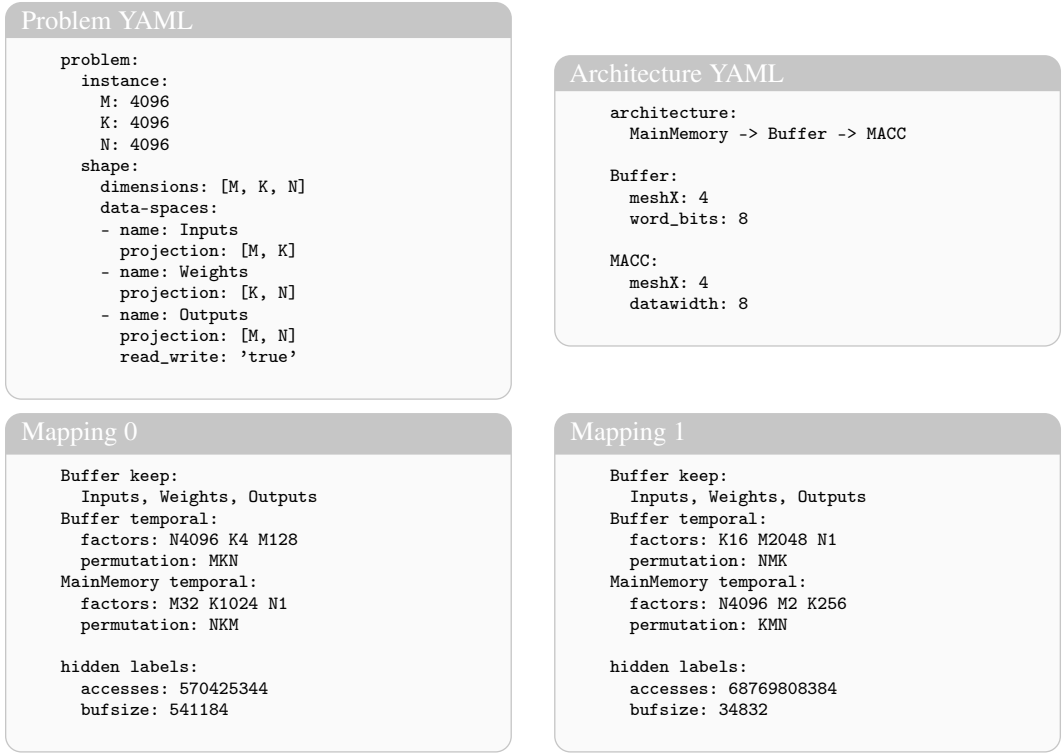

    \centering
    
    \begin{minipage}{0.48\linewidth}
    \begin{tcolorbox}[yamlbox,title=Problem YAML]
    \begin{verbatim}
    problem:
      instance:
        M: 4096
        K: 4096
        N: 4096
      shape:
        dimensions: [M, K, N]
        data-spaces:
        - name: Inputs
          projection: [M, K]
        - name: Weights
          projection: [K, N]
        - name: Outputs
          projection: [M, N]
          read_write: 'true'
    \end{verbatim}
    \end{tcolorbox}
    \end{minipage}
    \hfill
    \begin{minipage}{0.48\linewidth}
    \begin{tcolorbox}[yamlbox,title=Architecture YAML]
    \begin{verbatim}
    architecture:
      MainMemory -> Buffer -> MACC
    
    Buffer:
      meshX: 4
      word_bits: 8
    
    MACC:
      meshX: 4
      datawidth: 8
    \end{verbatim}
    \end{tcolorbox}
    \end{minipage}
    
    \vspace{0.4em}
    
    \begin{minipage}{0.48\linewidth}
    \begin{tcolorbox}[yamlbox,title=Mapping 0]
    \begin{verbatim}
    Buffer keep:
      Inputs, Weights, Outputs
    Buffer temporal:
      factors: N4096 K4 M128
      permutation: MKN
    MainMemory temporal:
      factors: M32 K1024 N1
      permutation: NKM
    
    hidden labels:
      accesses: 570425344
      bufsize: 541184
    \end{verbatim}
    \end{tcolorbox}
    \end{minipage}
    \hfill
    \begin{minipage}{0.48\linewidth}
    \begin{tcolorbox}[yamlbox,title=Mapping 1]
    \begin{verbatim}
    Buffer keep:
      Inputs, Weights, Outputs
    Buffer temporal:
      factors: K16 M2048 N1
      permutation: NMK
    MainMemory temporal:
      factors: N4096 M2 K256
      permutation: KMN
    
    hidden labels:
      accesses: 68769808384
      bufsize: 34832
    \end{verbatim}
    \end{tcolorbox}
    \end{minipage}
    
    
    \caption{
    Example benchmark case. The generated program receives the problem, architecture,
    and mapping YAML files, but not the hidden labels shown here. The two mappings
    implement the same $4096^3$ matrix multiplication but induce very different
    Timeloop-derived traffic and buffer requirements: mapping 0 has $120.6\times$
    lower \texttt{MainMemory} accesses, while mapping 1 uses $15.5\times$ less
    \texttt{Buffer} capacity.
    }
    \label{fig:benchmark-example-yaml}
    \end{figure}

    Figure~\ref{fig:benchmark-example-yaml} shows one complete benchmark case as the
    generated program sees it, together with the hidden ground truth.
    generated program sees it, together with the hidden reference labels.
    Table~\ref{tab:benchmark_components} summarizes the two components of the
    benchmark and what each is designed to test.
    
    \begin{table}[t]
    \centering
    \caption{Overview of the two benchmark components. QA-understanding tests qualitative mapping intuition, while performance modeling evaluates quantitative prediction ability.}
    \label{tab:benchmark_components}
    \small
    \begin{adjustbox}{max width=\linewidth}
    \begin{tabular}{p{0.17\linewidth} p{0.29\linewidth} p{0.27\linewidth} p{0.18\linewidth}}
    \toprule
    \textbf{Component} & \textbf{Task} & \textbf{Purpose} & \textbf{Stylized example} \\
    \midrule
    \textbf{Q\&A Mapping Understanding}
    &
    Given two mappings for the same tensor program, answer questions about their relative MainMemory access behavior.
    &
    Tests whether the model understands tiling, loop order, reuse, memory hierarchy placement, and keep annotations, independent of detailed numeric prediction.
    &
    \begin{minipage}[t]{\linewidth}
    \scriptsize
    \texttt{Map A: tile M,K}\\
    \texttt{Map B: tile N,K}\\
    Which has fewer MainMemory accesses?
    \end{minipage}
    \\
    \midrule
    \textbf{Analytical Performance Modeling}
    &
    Given a tensor program and mapping, predict quantitative performance-model quantities such as memory accesses, buffer sizes, or ratios.
    &
    Evaluates whether the model can translate mapping understanding into accurate quantitative estimates; errors may include conceptual, arithmetic, or calibration failures.
    &
    \begin{minipage}[t]{\linewidth}
    \footnotesize
    \texttt{MatMul: MK,KN$\to$MN}\\
    \texttt{Mapping: tile M=64,K=8}\\
    Predict MainMemory accesses.
    \end{minipage}
    \\
    \bottomrule
    \end{tabular}
    \end{adjustbox}
    \end{table}
    
    \section{Metric Definitions}
    \label{app:metrics}
    
    This appendix gives the formal definitions of the metrics summarized in
    Section~\ref{sec:methodology}.
    
    We use a strict exact-match metric (within a small error tolerance) as our main measure, i.e., the overall pass rate: a case passes only when the generated performance model predicts both total MainMemory accesses and required Buffer capacity correctly, up to a small numerical tolerance. This captures whether the synthesized model fully matches the Timeloop-derived analytical result, rather than only getting one part of the memory hierarchy right.
    
    We also report metric-specific exact rates for accesses and bufsize separately. Accesses exactness reflects whether the model captures data movement, reuse, bypass behavior, and loop-order effects, while bufsize exactness reflects whether it correctly computes the resident on-chip tile footprint. We use q-error to measure how far predictions are from ground truth on a multiplicative scale. Since errors can span many orders of magnitude, we use this multiplicative metric to differentiate between small and large errors in predictions more clearly than absolute error.
    We also report metric-specific exact rates for accesses and bufsize separately. Accesses exactness reflects whether the model captures data movement, reuse, bypass behavior, and loop-order effects, while bufsize exactness reflects whether it correctly computes the resident on-chip tile footprint. We use q-error to measure how far predictions are from the reference labels on a multiplicative scale. Since errors can span many orders of magnitude, we use this multiplicative metric to differentiate between small and large errors in predictions more clearly than absolute error.
    
    
    Let $y_{i,m}$ denote the Timeloop ground-truth value for case $i$ and metric
    Let $y_{i,m}$ denote the Timeloop reference value for case $i$ and metric
    $m \in \{\texttt{accesses}, \texttt{bufsize}\}$, and let $\hat{y}_{i,m}$
    denote the value predicted by the generated program. We treat a prediction as
    exact if it matches the ground truth within a small numerical tolerance, used
    exact if it matches the reference value within a small numerical tolerance, used
    only to avoid penalizing harmless integer-versus-floating-point formatting
    differences.
    The primary metric is both-exact pass rate:
    \[
    \mathrm{Pass}_{\mathrm{both}}
    =
    \frac{1}{N}
    \sum_{i=1}^{N}
    \mathbf{1}
    \left[
    \hat{y}_{i,\texttt{accesses}} \approx y_{i,\texttt{accesses}}
    \;\land\;
    \hat{y}_{i,\texttt{bufsize}} \approx y_{i,\texttt{bufsize}}
    \right],
    \]
    where $N=216$ in our benchmark. This metric is intentionally strict: a case is
    correct only if the model simultaneously predicts off-chip traffic and on-chip
    storage correctly.
    We also report per-metric exact pass rates:
    \[
    \mathrm{Pass}_{m}
    =
    \frac{1}{N}
    \sum_{i=1}^{N}
    \mathbf{1}
    \left[
    \hat{y}_{i,m} \approx y_{i,m}
    \right],
    \qquad
    m \in \{\texttt{accesses}, \texttt{bufsize}\}.
    \]
    
    In addition to accuracy and pass rates, to better understand how off the model's estimates are, we use a symmetric multiplicative error or q-error \cite{qerror}:
    \[
    Q(\hat{y}, y)
    =
    \max
    \left(
    \frac{\max(\hat{y},1)}{\max(y,1)},
    \frac{\max(y,1)}{\max(\hat{y},1)}
    \right).
    \]
    A q-error of $1$ is exact, a q-error of $2$ means the estimate is off by a factor of two a q-error of $10$ indicates the prediction is off from the ground truth by ten times, etc. Larger values indicate larger multiplicative error, potentially in either direction. The q-error provides additional information that overall pass rates and accuracy do not: for example, in the case where two models are both wrong for a given test case, one model can be worse at the prediction than the other (e.g., estimates that are twice off versus one that is $10^8$ times off).


    Finally, as each problem has two mappings, we also evaluate whether a generated performance model preserves the relative ordering of mappings by memory traffic so as to evaluate not just exact correctness but also its ability to rank mappings via said program. Let
    $\mathcal{P}$ be the set of mapping pairs with unequal true access counts. For
    a pair $(a,b) \in \mathcal{P}$, let $A_a$ and $A_b$ be the ground-truth
    \texttt{accesses}, and let $\hat{A}_a$ and $\hat{A}_b$ be the predicted
    \texttt{accesses}. Pairwise preference accuracy is:
    \[
    \mathrm{PrefAcc}
    =
    \frac{1}{|\mathcal{P}|}
    \sum_{(a,b)\in\mathcal{P}}
    \mathbf{1}
    \left[
    (\hat{A}_a < \hat{A}_b) = (A_a < A_b)
    \right].
    \]
    This measures whether the generated model can identify the lower-traffic
    mapping, even when absolute traffic estimates are imperfect.


    \section{Detailed Q\&A Analysis}
    
    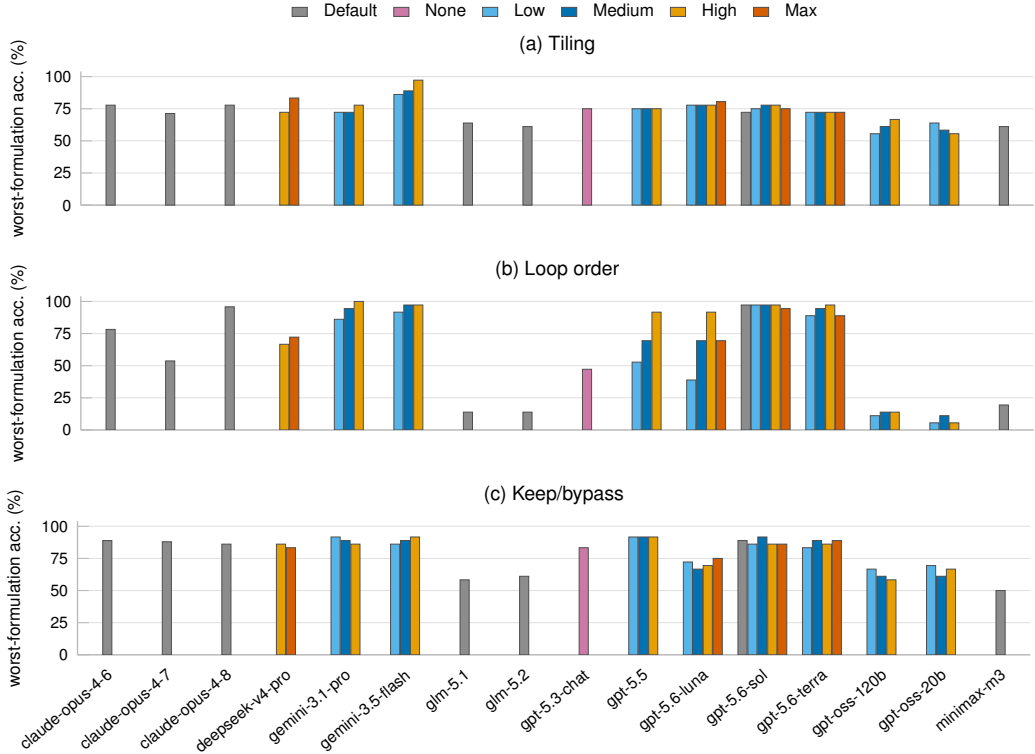
\begin{figure}[ht]
      \centering
\begin{tikzpicture}[font=\sffamily]
\begin{axis}[
  x=0.1313cm, height=3.35cm,
  xmin=-3.00, xmax=93.00, ymin=0, ymax=104,
  ytick={0,25,50,75,100},
  ylabel={worst-formulation acc.\ (\%)},
  ylabel style={font=\sffamily\fontsize{7}{8}\selectfont},
  title={(a) Tiling},
  title style={font=\sffamily\fontsize{8}{9}\selectfont, yshift=-3pt},
  xtick={0.000,6.000,12.000,18.000,24.000,30.000,36.000,42.000,48.000,54.000,60.000,66.000,72.000,78.000,84.000,90.000},
  /pgf/number format/assume math mode=true,
  ymajorgrids, grid style={gray!25, line width=0.3pt},
  axis line style={gray!55, line width=0.5pt}, tick style={gray!55},
  axis x line*=bottom, axis y line*=left,
  y tick label style={font=\sffamily\fontsize{6.5}{8}\selectfont},
  xticklabels={},
  legend style={at={(0.5,1.30)}, anchor=south, legend columns=6, draw=none,
                fill=none, font=\sffamily\fontsize{7}{8}\selectfont, column sep=0.9ex},
  legend image code/.code={\draw[#1] (0cm,-0.06cm) rectangle (0.20cm,0.06cm);},
]
\addplot[ybar, bar shift=0pt, bar width=3.44pt, fill=black!45, draw=black!70, line width=0.25pt] coordinates {(0.000,77.778) (6.000,71.296) (12.000,77.778) (36.000,63.889) (42.000,61.111) (64.000,72.222) (90.000,61.111)};
\addlegendentry{Default}
\addplot[ybar, bar shift=0pt, bar width=3.44pt, fill=okpurple, draw=black!70, line width=0.25pt] coordinates {(48.000,75.000)};
\addlegendentry{None}
\addplot[ybar, bar shift=0pt, bar width=3.44pt, fill=okcyan, draw=black!70, line width=0.25pt] coordinates {(23.000,72.222) (29.000,86.111) (53.000,75.000) (58.500,77.778) (65.000,75.000) (70.500,72.222) (77.000,55.556) (83.000,63.889)};
\addlegendentry{Low}
\addplot[ybar, bar shift=0pt, bar width=3.44pt, fill=okblue, draw=black!70, line width=0.25pt] coordinates {(24.000,72.222) (30.000,88.889) (54.000,75.000) (59.500,77.778) (66.000,77.778) (71.500,72.222) (78.000,61.111) (84.000,58.333)};
\addlegendentry{Medium}
\addplot[ybar, bar shift=0pt, bar width=3.44pt, fill=okorange, draw=black!70, line width=0.25pt] coordinates {(17.500,72.222) (25.000,77.778) (31.000,97.222) (55.000,75.000) (60.500,77.778) (67.000,77.778) (72.500,72.222) (79.000,66.667) (85.000,55.556)};
\addlegendentry{High}
\addplot[ybar, bar shift=0pt, bar width=3.44pt, fill=okred, draw=black!70, line width=0.25pt] coordinates {(18.500,83.333) (61.500,80.556) (68.000,75.000) (73.500,72.222)};
\addlegendentry{Max}
\end{axis}
\end{tikzpicture}
\\[1pt]
\begin{tikzpicture}[font=\sffamily]
\begin{axis}[
  x=0.1313cm, height=3.35cm,
  xmin=-3.00, xmax=93.00, ymin=0, ymax=104,
  ytick={0,25,50,75,100},
  ylabel={worst-formulation acc.\ (\%)},
  ylabel style={font=\sffamily\fontsize{7}{8}\selectfont},
  title={(b) Loop order},
  title style={font=\sffamily\fontsize{8}{9}\selectfont, yshift=-3pt},
  xtick={0.000,6.000,12.000,18.000,24.000,30.000,36.000,42.000,48.000,54.000,60.000,66.000,72.000,78.000,84.000,90.000},
  /pgf/number format/assume math mode=true,
  ymajorgrids, grid style={gray!25, line width=0.3pt},
  axis line style={gray!55, line width=0.5pt}, tick style={gray!55},
  axis x line*=bottom, axis y line*=left,
  y tick label style={font=\sffamily\fontsize{6.5}{8}\selectfont},
  xticklabels={},
]
\addplot[ybar, bar shift=0pt, bar width=3.44pt, fill=black!45, draw=black!70, line width=0.25pt, forget plot] coordinates {(0.000,78.241) (6.000,53.704) (12.000,95.833) (36.000,13.889) (42.000,13.889) (64.000,97.222) (90.000,19.444)};
\addplot[ybar, bar shift=0pt, bar width=3.44pt, fill=okpurple, draw=black!70, line width=0.25pt, forget plot] coordinates {(48.000,47.222)};
\addplot[ybar, bar shift=0pt, bar width=3.44pt, fill=okcyan, draw=black!70, line width=0.25pt, forget plot] coordinates {(23.000,86.111) (29.000,91.667) (53.000,52.778) (58.500,38.889) (65.000,97.222) (70.500,88.889) (77.000,11.111) (83.000,5.556)};
\addplot[ybar, bar shift=0pt, bar width=3.44pt, fill=okblue, draw=black!70, line width=0.25pt, forget plot] coordinates {(24.000,94.444) (30.000,97.222) (54.000,69.444) (59.500,69.444) (66.000,97.222) (71.500,94.444) (78.000,13.889) (84.000,11.111)};
\addplot[ybar, bar shift=0pt, bar width=3.44pt, fill=okorange, draw=black!70, line width=0.25pt, forget plot] coordinates {(17.500,66.667) (25.000,100.000) (31.000,97.222) (55.000,91.667) (60.500,91.667) (67.000,97.222) (72.500,97.222) (79.000,13.889) (85.000,5.556)};
\addplot[ybar, bar shift=0pt, bar width=3.44pt, fill=okred, draw=black!70, line width=0.25pt, forget plot] coordinates {(18.500,72.222) (61.500,69.444) (68.000,94.444) (73.500,88.889)};
\end{axis}
\end{tikzpicture}
\\[1pt]
\begin{tikzpicture}[font=\sffamily]
\begin{axis}[
  x=0.1313cm, height=3.35cm,
  xmin=-3.00, xmax=93.00, ymin=0, ymax=104,
  ytick={0,25,50,75,100},
  ylabel={worst-formulation acc.\ (\%)},
  ylabel style={font=\sffamily\fontsize{7}{8}\selectfont},
  title={(c) Keep/bypass},
  title style={font=\sffamily\fontsize{8}{9}\selectfont, yshift=-3pt},
  xtick={0.000,6.000,12.000,18.000,24.000,30.000,36.000,42.000,48.000,54.000,60.000,66.000,72.000,78.000,84.000,90.000},
  /pgf/number format/assume math mode=true,
  ymajorgrids, grid style={gray!25, line width=0.3pt},
  axis line style={gray!55, line width=0.5pt}, tick style={gray!55},
  axis x line*=bottom, axis y line*=left,
  y tick label style={font=\sffamily\fontsize{6.5}{8}\selectfont},
  xticklabels={{claude-opus-4-6},{claude-opus-4-7},{claude-opus-4-8},{deepseek-v4-pro},{gemini-3.1-pro},{gemini-3.5-flash},{glm-5.1},{glm-5.2},{gpt-5.3-chat},{gpt-5.5},{gpt-5.6-luna},{gpt-5.6-sol},{gpt-5.6-terra},{gpt-oss-120b},{gpt-oss-20b},{minimax-m3}},
  x tick label style={font=\sffamily\fontsize{6.5}{8}\selectfont, rotate=40, anchor=north east},
]
\addplot[ybar, bar shift=0pt, bar width=3.44pt, fill=black!45, draw=black!70, line width=0.25pt, forget plot] coordinates {(0.000,88.889) (6.000,87.963) (12.000,86.111) (36.000,58.333) (42.000,61.111) (64.000,88.889) (90.000,50.000)};
\addplot[ybar, bar shift=0pt, bar width=3.44pt, fill=okpurple, draw=black!70, line width=0.25pt, forget plot] coordinates {(48.000,83.333)};
\addplot[ybar, bar shift=0pt, bar width=3.44pt, fill=okcyan, draw=black!70, line width=0.25pt, forget plot] coordinates {(23.000,91.667) (29.000,86.111) (53.000,91.667) (58.500,72.222) (65.000,86.111) (70.500,83.333) (77.000,66.667) (83.000,69.444)};
\addplot[ybar, bar shift=0pt, bar width=3.44pt, fill=okblue, draw=black!70, line width=0.25pt, forget plot] coordinates {(24.000,88.889) (30.000,88.889) (54.000,91.667) (59.500,66.667) (66.000,91.667) (71.500,88.889) (78.000,61.111) (84.000,61.111)};
\addplot[ybar, bar shift=0pt, bar width=3.44pt, fill=okorange, draw=black!70, line width=0.25pt, forget plot] coordinates {(17.500,86.111) (25.000,86.111) (31.000,91.667) (55.000,91.667) (60.500,69.444) (67.000,86.111) (72.500,86.111) (79.000,58.333) (85.000,66.667)};
\addplot[ybar, bar shift=0pt, bar width=3.44pt, fill=okred, draw=black!70, line width=0.25pt, forget plot] coordinates {(18.500,83.333) (61.500,75.000) (68.000,86.111) (73.500,88.889)};
\end{axis}
\end{tikzpicture}
      \caption{Worst-formulation Q\&A accuracy by mapping axis. Loop-order
      sensitivity most clearly separates strong and weak configurations.}
      \label{fig:qa-axis-4p}
    \end{figure}
    
    \begin{figure}[ht]
      \centering
\begin{tikzpicture}[font=\sffamily, baseline=(current bounding box.north)]
\begin{axis}[
  width=0.50\linewidth, height=4.4cm,
  xlabel style={font=\sffamily\fontsize{7}{8}\selectfont},
  ylabel style={font=\sffamily\fontsize{7}{8}\selectfont},
  tick label style={font=\sffamily\fontsize{6.5}{8}\selectfont},
  /pgf/number format/assume math mode=true,
  ymajorgrids, grid style={gray!25, line width=0.3pt},
  axis line style={gray!55, line width=0.5pt}, tick style={gray!55},
  axis x line*=bottom, axis y line*=left,
  title style={font=\sffamily\fontsize{7.5}{9}\selectfont, yshift=-2pt},
  title={(a) permutation-blindness predicts overall skill},
  xlabel={``both equal'' answers on loop-order pairs (\%)},
  ylabel={worst-formulation acc.\ (\%)},
  xmin=-4, xmax=90, ymin=20, ymax=100,
]
\addplot[only marks, mark=*, mark size=1.15pt, draw=okblue, fill=okblue] coordinates {(77.778,43.519) (77.778,43.519) (75.000,45.370) (83.333,46.296) (66.667,48.148) (61.111,48.148) (58.333,50.000) (77.778,50.000) (66.667,51.852) (36.111,64.815) (36.111,70.370) (13.889,71.296) (25.000,72.222) (41.667,73.148) (13.889,76.852) (16.667,78.704) (13.889,79.630) (2.778,80.556) (8.333,81.481) (18.519,82.407) (19.444,82.407) (8.333,84.259) (2.778,85.185) (2.778,85.185) (8.333,86.111) (0.000,87.037) (0.000,87.037) (0.000,87.037) (5.556,87.037) (2.778,87.500) (2.778,87.963) (2.778,88.889) (0.000,89.815) (0.000,89.815) (0.000,89.815) (2.778,92.593) (0.000,95.370)};
\addplot[okred, line width=0.7pt, dashed, forget plot] coordinates {(0,88.555) (83.333,41.481)};
\node[anchor=north east, font=\sffamily\fontsize{7}{8}\selectfont, text=okred] at (rel axis cs:0.98,0.97) {$r=-0.97$};
\end{axis}
\end{tikzpicture}%
\hfill%
\begin{tikzpicture}[font=\sffamily, baseline=(current bounding box.north)]
\begin{axis}[
  width=0.50\linewidth, height=4.4cm,
  xlabel style={font=\sffamily\fontsize{7}{8}\selectfont},
  ylabel style={font=\sffamily\fontsize{7}{8}\selectfont},
  tick label style={font=\sffamily\fontsize{6.5}{8}\selectfont},
  /pgf/number format/assume math mode=true,
  ymajorgrids, grid style={gray!25, line width=0.3pt},
  axis line style={gray!55, line width=0.5pt}, tick style={gray!55},
  axis x line*=bottom, axis y line*=left,
  title style={font=\sffamily\fontsize{7.5}{9}\selectfont, yshift=-2pt},
  title={(b) difficulty tracks how far apart the mappings are},
  xlabel={ground-truth MainMemory access ratio between the pair},
  ylabel={worst-formulation acc.\ (\%)},
  symbolic x coords={$<$1.05$\times$,1.05--1.5$\times$,1.5--4$\times$,4--20$\times$,$>$20$\times$},
  xtick=data, ymin=40, ymax=100,
  x tick label style={font=\sffamily\fontsize{6}{7}\selectfont, rotate=25, anchor=north east},
]
\addplot[okblue, line width=0.9pt, mark=*, mark size=1.4pt, mark options={fill=okblue}] coordinates {($<$1.05$\times$,67.160) (1.05--1.5$\times$,71.538) (1.5--4$\times$,71.515) (4--20$\times$,80.889) ($>$20$\times$,97.037)};
\end{axis}
\end{tikzpicture}
      \caption{Loop-order permutation blindness: ``equal'' responses strongly
      predict loop-order errors and overall Q\&A accuracy.}
      \label{fig:qa-difficulty-4p}
    \end{figure}
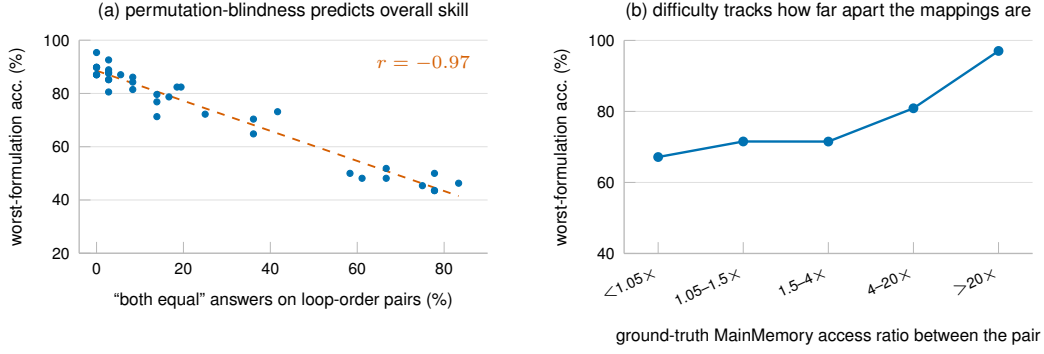
    
    \begin{figure}[ht]
      \centering
\begin{tikzpicture}[font=\sffamily]
\begin{axis}[
  x=0.1313cm, height=3.35cm,
  xmin=-3.00, xmax=93.00, ymin=0, ymax=104,
  ytick={0,25,50,75,100},
  ylabel={accuracy (\%)},
  ylabel style={font=\sffamily\fontsize{7}{8}\selectfont},
  title={(a) MatMul},
  title style={font=\sffamily\fontsize{8}{9}\selectfont, yshift=-3pt},
  xtick={0.000,6.000,12.000,18.000,24.000,30.000,36.000,42.000,48.000,54.000,60.000,66.000,72.000,78.000,84.000,90.000},
  ymajorgrids, grid style={gray!25, line width=0.3pt},
  axis line style={gray!55, line width=0.5pt}, tick style={gray!55},
  axis x line*=bottom, axis y line*=left,
  y tick label style={font=\sffamily\fontsize{6.5}{8}\selectfont},
  xticklabels={},
  legend style={at={(0.5,1.30)}, anchor=south, legend columns=6, draw=none,
                fill=none, font=\sffamily\fontsize{7}{8}\selectfont, column sep=0.9ex},
  legend image code/.code={\draw[#1] (0cm,-0.06cm) rectangle (0.20cm,0.06cm);},
]
\addplot[ybar, bar shift=0pt, bar width=3.44pt, fill=black!45, draw=black!70, line width=0.25pt] coordinates {(0.000,72.222) (6.000,62.037) (12.000,79.167) (36.000,33.333) (42.000,25.000) (64.000,77.778) (90.000,30.556)};
\addlegendentry{Default}
\addplot[ybar, bar shift=0pt, bar width=3.44pt, fill=okpurple, draw=black!70, line width=0.25pt] coordinates {(48.000,50.000)};
\addlegendentry{None}
\addplot[ybar, bar shift=0pt, bar width=3.44pt, fill=okcyan, draw=black!70, line width=0.25pt] coordinates {(23.000,75.000) (29.000,83.333) (53.000,66.667) (58.500,47.222) (65.000,80.556) (70.500,72.222) (77.000,33.333) (83.000,33.333)};
\addlegendentry{Low}
\addplot[ybar, bar shift=0pt, bar width=3.44pt, fill=okblue, draw=black!70, line width=0.25pt] coordinates {(24.000,86.111) (30.000,86.111) (54.000,83.333) (59.500,58.333) (66.000,86.111) (71.500,80.556) (78.000,22.222) (84.000,33.333)};
\addlegendentry{Medium}
\addplot[ybar, bar shift=0pt, bar width=3.44pt, fill=okorange, draw=black!70, line width=0.25pt] coordinates {(17.500,63.889) (25.000,83.333) (31.000,91.667) (55.000,86.111) (60.500,75.000) (67.000,77.778) (72.500,77.778) (79.000,38.889) (85.000,27.778)};
\addlegendentry{High}
\addplot[ybar, bar shift=0pt, bar width=3.44pt, fill=okred, draw=black!70, line width=0.25pt] coordinates {(18.500,66.667) (61.500,66.667) (68.000,80.556) (73.500,77.778)};
\addlegendentry{Max}
\end{axis}
\end{tikzpicture}
\\[1pt]
\begin{tikzpicture}[font=\sffamily]
\begin{axis}[
  x=0.1313cm, height=3.35cm,
  xmin=-3.00, xmax=93.00, ymin=0, ymax=104,
  ytick={0,25,50,75,100},
  ylabel={accuracy (\%)},
  ylabel style={font=\sffamily\fontsize{7}{8}\selectfont},
  title={(b) Batched MatMul},
  title style={font=\sffamily\fontsize{8}{9}\selectfont, yshift=-3pt},
  xtick={0.000,6.000,12.000,18.000,24.000,30.000,36.000,42.000,48.000,54.000,60.000,66.000,72.000,78.000,84.000,90.000},
  ymajorgrids, grid style={gray!25, line width=0.3pt},
  axis line style={gray!55, line width=0.5pt}, tick style={gray!55},
  axis x line*=bottom, axis y line*=left,
  y tick label style={font=\sffamily\fontsize{6.5}{8}\selectfont},
  xticklabels={},
]
\addplot[ybar, bar shift=0pt, bar width=3.44pt, fill=black!45, draw=black!70, line width=0.25pt, forget plot] coordinates {(0.000,71.296) (6.000,53.704) (12.000,69.444) (36.000,16.667) (42.000,25.000) (64.000,75.000) (90.000,19.444)};
\addplot[ybar, bar shift=0pt, bar width=3.44pt, fill=okpurple, draw=black!70, line width=0.25pt, forget plot] coordinates {(48.000,50.000)};
\addplot[ybar, bar shift=0pt, bar width=3.44pt, fill=okcyan, draw=black!70, line width=0.25pt, forget plot] coordinates {(23.000,66.667) (29.000,83.333) (53.000,47.222) (58.500,36.111) (65.000,77.778) (70.500,63.889) (77.000,33.333) (83.000,16.667)};
\addplot[ybar, bar shift=0pt, bar width=3.44pt, fill=okblue, draw=black!70, line width=0.25pt, forget plot] coordinates {(24.000,75.000) (30.000,91.667) (54.000,55.556) (59.500,55.556) (66.000,77.778) (71.500,69.444) (78.000,22.222) (84.000,22.222)};
\addplot[ybar, bar shift=0pt, bar width=3.44pt, fill=okorange, draw=black!70, line width=0.25pt, forget plot] coordinates {(17.500,63.889) (25.000,80.556) (31.000,91.667) (55.000,75.000) (60.500,63.889) (67.000,77.778) (72.500,77.778) (79.000,27.778) (85.000,25.000)};
\addplot[ybar, bar shift=0pt, bar width=3.44pt, fill=okred, draw=black!70, line width=0.25pt, forget plot] coordinates {(18.500,61.111) (61.500,55.556) (68.000,72.222) (73.500,66.667)};
\end{axis}
\end{tikzpicture}
\\[1pt]
\begin{tikzpicture}[font=\sffamily]
\begin{axis}[
  x=0.1313cm, height=3.35cm,
  xmin=-3.00, xmax=93.00, ymin=0, ymax=104,
  ytick={0,25,50,75,100},
  ylabel={accuracy (\%)},
  ylabel style={font=\sffamily\fontsize{7}{8}\selectfont},
  title={(c) Conv2D},
  title style={font=\sffamily\fontsize{8}{9}\selectfont, yshift=-3pt},
  xtick={0.000,6.000,12.000,18.000,24.000,30.000,36.000,42.000,48.000,54.000,60.000,66.000,72.000,78.000,84.000,90.000},
  ymajorgrids, grid style={gray!25, line width=0.3pt},
  axis line style={gray!55, line width=0.5pt}, tick style={gray!55},
  axis x line*=bottom, axis y line*=left,
  y tick label style={font=\sffamily\fontsize{6.5}{8}\selectfont},
  xticklabels={{claude-opus-4-6},{claude-opus-4-7},{claude-opus-4-8},{deepseek-v4-pro},{gemini-3.1-pro},{gemini-3.5-flash},{glm-5.1},{glm-5.2},{gpt-5.3-chat},{gpt-5.5},{gpt-5.6-luna},{gpt-5.6-sol},{gpt-5.6-terra},{gpt-oss-120b},{gpt-oss-20b},{minimax-m3}},
  x tick label style={font=\sffamily\fontsize{6.5}{8}\selectfont, rotate=40, anchor=north east},
]
\addplot[ybar, bar shift=0pt, bar width=3.44pt, fill=black!45, draw=black!70, line width=0.25pt, forget plot] coordinates {(0.000,75.463) (6.000,62.037) (12.000,81.944) (36.000,38.889) (42.000,36.111) (64.000,100.000) (90.000,38.889)};
\addplot[ybar, bar shift=0pt, bar width=3.44pt, fill=okpurple, draw=black!70, line width=0.25pt, forget plot] coordinates {(48.000,61.111)};
\addplot[ybar, bar shift=0pt, bar width=3.44pt, fill=okcyan, draw=black!70, line width=0.25pt, forget plot] coordinates {(23.000,77.778) (29.000,80.556) (53.000,80.556) (58.500,55.556) (65.000,97.222) (70.500,91.667) (77.000,36.111) (83.000,41.667)};
\addplot[ybar, bar shift=0pt, bar width=3.44pt, fill=okblue, draw=black!70, line width=0.25pt, forget plot] coordinates {(24.000,94.444) (30.000,91.667) (54.000,88.889) (59.500,47.222) (66.000,100.000) (71.500,94.444) (78.000,44.444) (84.000,38.889)};
\addplot[ybar, bar shift=0pt, bar width=3.44pt, fill=okorange, draw=black!70, line width=0.25pt, forget plot] coordinates {(17.500,58.333) (25.000,100.000) (31.000,94.444) (55.000,97.222) (60.500,80.556) (67.000,100.000) (72.500,97.222) (79.000,50.000) (85.000,41.667)};
\addplot[ybar, bar shift=0pt, bar width=3.44pt, fill=okred, draw=black!70, line width=0.25pt, forget plot] coordinates {(18.500,75.000) (61.500,69.444) (68.000,100.000) (73.500,91.667)};
\end{axis}
\end{tikzpicture}
      \caption{Worst-formulation Q\&A accuracy by workload family.}
      \label{fig:qa-workload-4p}
    \end{figure}
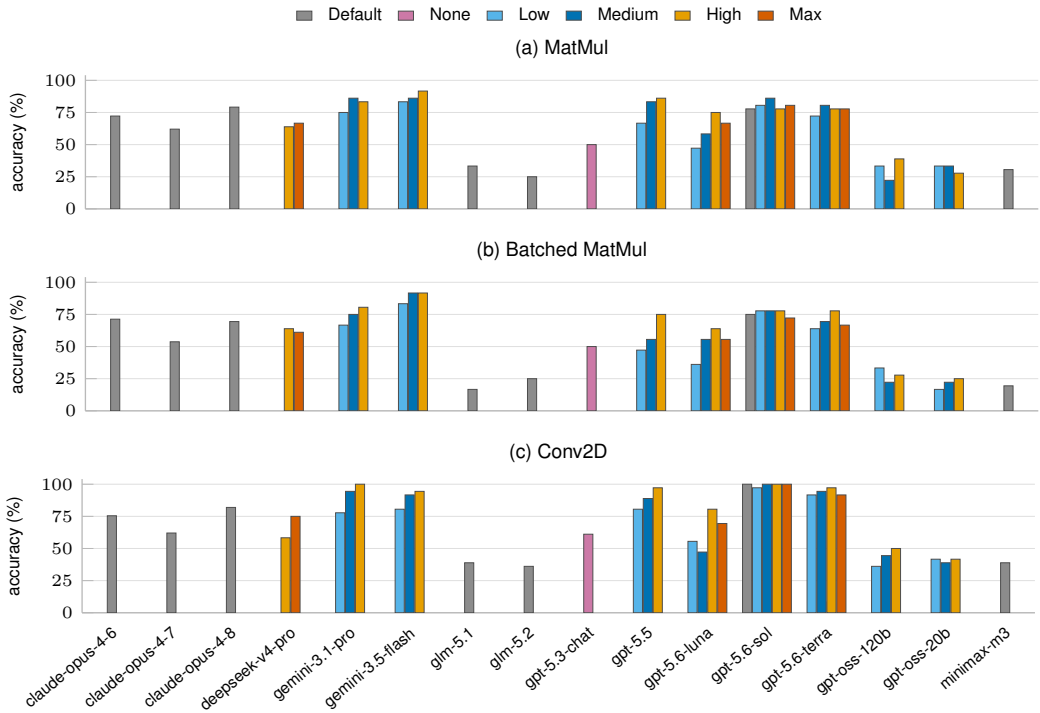
    
    \begin{table}[t]
    \centering
    \caption{Pooled all-three Q\&A accuracy (\%) for each cell of the balanced
    $3\times3$ benchmark design (12 problems per cell).}
    \label{tab:qa-opt-by-workload-4p}
    \small
    \begin{tabular}{@{}l rrr@{}}
    \toprule
    Distinguishing axis & MatMul & Batched MM & Conv2D \\
    \midrule
    Tiling & 72.8 & 64.8 & 88.5 \\
    Loop order & 71.5 & 68.0 & 69.3 \\
    Bypass & 83.0 & 78.3 & 87.4 \\
    \bottomrule
    \end{tabular}
    \end{table}
    
    Loop order is bimodal across model families: configurations either track the
    reuse consequences of permutation or frequently declare the mappings equal.
    Tiling and tensor retention vary more smoothly. Convolution is generally the
    easiest workload and batched matrix multiplication the hardest, but the mapping
    axis explains more of the cross-model separation than workload family.
    
    \section{Progress and Domain-Specific Fine-Tuning}
    
    \subsection{Progress Across Model Generations}
    \label{subsec:qa-history}
    
    \begin{tcolorbox}[colback=blue!5,colframe=blue!40!black]
    \textbf{Key Takeaway:} Closed models now achieve near-perfect QA accuracy
    ($95.4\%$), while the best open-weight model reaches $82.4\%$; both improve
    substantially over the 2025 snapshot, although a 13-point gap remains.
    \end{tcolorbox}
    
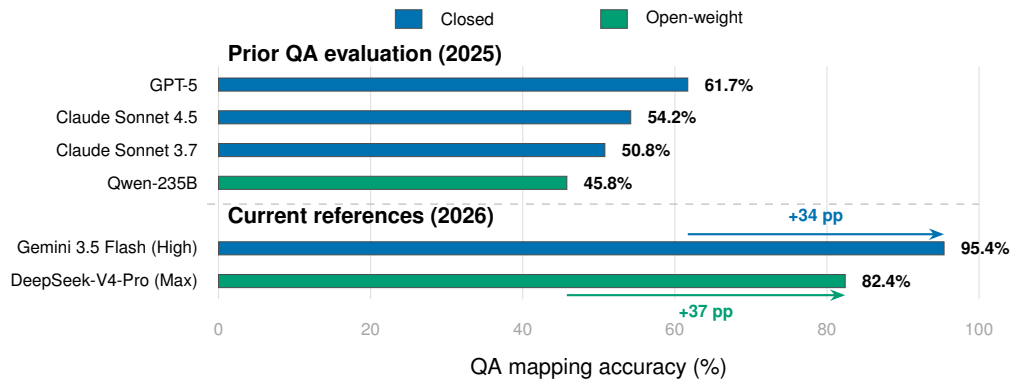
\begin{figure}[t]
        \centering
        \begin{tikzpicture}[x=0.0072\linewidth,y=0.4333cm,font=\sffamily\scriptsize]
            \foreach \score in {0,20,40,60,80,100} {
                \draw[gray!25] (\score,-1.55) -- (\score,5.55);
                \node[anchor=north,text=gray!70] at (\score,-2.00) {\score};
            }
            \node[font=\sffamily\small] at (50,-3.65) {QA mapping accuracy (\%)};
    
            \newcommand{\historybar}[4]{%
                \filldraw[fill=#3, draw=black!70, line width=0.25pt]
                    (0,#1-0.20) rectangle (#2,#1+0.20);
                \node[anchor=east] at (-1.5,#1) {#4};
                \node[anchor=west,font=\sffamily\scriptsize\bfseries] at (#2+1,#1) {#2\%};
            }
    
            \node[anchor=west,font=\sffamily\small\bfseries] at (0,5.9)
                {Prior QA evaluation (2025)};
            \historybar{5}{61.7}{okblue}{GPT-5}
            \historybar{4}{54.2}{okblue}{Claude Sonnet 4.5}
            \historybar{3}{50.8}{okblue}{Claude Sonnet 3.7}
            \historybar{2}{45.8}{okgreen}{Qwen-235B}
    
            \draw[gray!60,dashed] (-1.5,1.35) -- (100,1.35);
            \node[anchor=west,font=\sffamily\small\bfseries] at (0,0.95)
                {Current references (2026)};
            \historybar{0}{95.4}{okblue}{Gemini 3.5 Flash (High)}
            \historybar{-1}{82.4}{okgreen}{DeepSeek-V4-Pro (Max)}
    
            \draw[-{Stealth[length=2mm]},thick,okblue]
                (61.7,0.43) -- node[above,font=\sffamily\scriptsize\bfseries] {+34 pp} (95.4,0.43);
            \draw[-{Stealth[length=2mm]},thick,okgreen]
                (45.8,-1.43) -- node[below,font=\sffamily\scriptsize\bfseries] {+37 pp} (82.4,-1.43);
    
            \node[fill=okblue,draw=black!70,line width=0.25pt,minimum width=1em,minimum height=0.7em,inner sep=0pt]
                at (25,7.0) {};
            \node[anchor=west] at (28,7.0) {Closed};
            \node[fill=okgreen,draw=black!70,line width=0.25pt,minimum width=1em,minimum height=0.7em,inner sep=0pt]
                at (52,7.0) {};
            \node[anchor=west] at (55,7.0) {Open-weight};
        \end{tikzpicture}
        \caption{QA mapping-understanding results from the prior cohort and the
        strongest current references in Figure~\ref{fig:qa_overall}. Prior values
        use direct-comparison accuracy on 120 examples; current values use the
        stricter worst-formulation metric. We treat the closely aligned tasks as roughly
        comparable; the reported gains are therefore approximate.}
        \label{fig:qa-progress}
    \end{figure}

    To place the current results in context, we compare them with our prior QA
    evaluation cohort. The earlier evaluation used 120 direct mapping-comparison
    examples. GPT-5 (August 2025) answered 74 correctly ($61.7\%$), followed by
    Claude Sonnet 4.5 (September 2025) at $54.2\%$, Claude Sonnet 3.7 (February
    2025) at $50.8\%$, and the open-weight Qwen3-235B-A22B (April 2025) at
    $45.8\%$. Figure~\ref{fig:qa-progress} contrasts these results with the
    strongest current references in Figure~\ref{fig:qa_overall}: Gemini 3.5 Flash
    (2026) reaches $95.4\%$ at high reasoning effort, while
    DeepSeek-V4-Pro (April 2026) reaches $82.4\%$ at max effort.
    
    The strongest closed-model reference rises by approximately 34 percentage
    points, while the strongest general-purpose open-weight reference rises by
    approximately 37 points. Because the cohorts test closely aligned
    mapping-comparison tasks, we treat them as roughly comparable snapshots. The
    current worst-formulation metric is stricter, however, so these gains are approximate
    rather than a controlled longitudinal estimate.
    
    \subsection{RL Fine-Tuning Can Improve Mapping Reasoning}
    \label{subsec:qa-rl}
    
    \begin{tcolorbox}[colback=blue!5,colframe=blue!40!black]
    \textbf{Key Takeaway:} Additional task-specific training helps: verifier-guided
    RL on binary Q\&A labels lifts a Qwen3-4B backbone from $54.3\%$ to $70\%$,
    overtaking GPT-5 (August 2025) at $61.7\%$. Domain-specific fine-tuning can
    inject mapping expertise into small open-weight models.
    \end{tcolorbox}
    
\begin{figure}[t]
        \centering
        \begin{tikzpicture}[x=0.0072\linewidth,y=0.4333cm,font=\sffamily\scriptsize]
            \foreach \score in {0,20,40,60,80,100} {
                \draw[gray!25] (\score,-0.70) -- (\score,2.55);
                \node[anchor=north,text=gray!70] at (\score,-1.15) {\score};
            }
            \node[font=\sffamily\small] at (50,-2.80) {QA mapping accuracy (\%)};
    
            \newcommand{\rlbar}[4]{%
                \filldraw[fill=#3, draw=black!70, line width=0.25pt]
                    (0,#1-0.20) rectangle (#2,#1+0.20);
                \node[anchor=east] at (-1.5,#1) {#4};
                \node[anchor=west,font=\sffamily\scriptsize\bfseries] at (#2+1,#1) {#2\%};
            }
    
            \rlbar{2}{54.3}{okgreen}{Qwen3-4B (base)}
            \rlbar{1}{61.7}{okblue}{GPT-5 (Aug 2025)}
            \rlbar{0}{70.0}{okorange}{Qwen3-4B + RL (ours)}
    
            \draw[-{Stealth[length=2mm]},thick,okorange]
                (54.3,-0.57) -- node[below,font=\sffamily\scriptsize\bfseries] {+15.7 pp} (70.0,-0.57);
    
            \node[fill=okgreen,draw=black!70,line width=0.25pt,minimum width=1em,minimum height=0.7em,inner sep=0pt]
                at (14,3.9) {};
            \node[anchor=west] at (17,3.9) {Open-weight};
            \node[fill=okblue,draw=black!70,line width=0.25pt,minimum width=1em,minimum height=0.7em,inner sep=0pt]
                at (44,3.9) {};
            \node[anchor=west] at (47,3.9) {Closed};
            \node[fill=okorange,draw=black!70,line width=0.25pt,minimum width=1em,minimum height=0.7em,inner sep=0pt]
                at (64,3.9) {};
            \node[anchor=west] at (67,3.9) {Ours (RL)};
        \end{tikzpicture}
        \caption{Verifier-guided RL fine-tuning on binary Q\&A labels
        lifts a Qwen3-4B backbone from $54.3\%$ to $70\%$ on the 120-example QA
        evaluation, past GPT-5 (August 2025) at $61.7\%$. All three use the prior
        cohort's direct-comparison metric and are therefore directly comparable.}
        \label{fig:qa-rl}
    \end{figure}
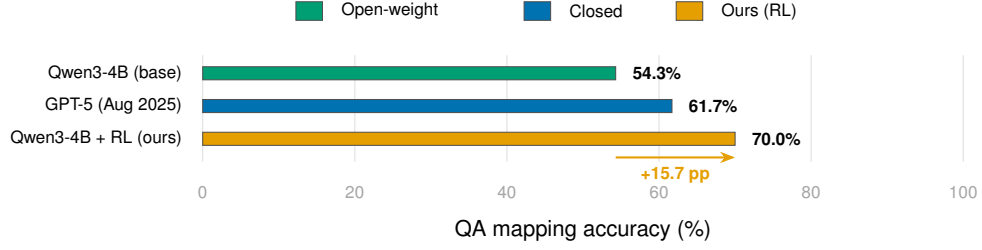

    The prior cohort also included our RL-fine-tuned Qwen3-4B (April 2025),
    shown in Figure~\ref{fig:qa-rl}. Verifier-guided RL on binary Q\&A labels lifts
    the same backbone from $54.3\%$ to $70\%$ on the 120-example evaluation, past
    GPT-5 (August 2025) by 8.3 points and the far larger Qwen3-235B-A22B by 24.2.
    Mapping reasoning suits this setup: Timeloop-derived labels give objective
    rewards, and diverse loop orders, tiling factors, and keep/bypass choices expose
    exactly the reuse patterns on which general-purpose models fail.
    
    The baseline is the same backbone without RL, so the $15.7$-point gain is
    measured against a matched control, though from a single pass over the smaller
    legacy suite. The study also predates this evaluation by roughly nine months,
    and both closed-source and open-weight models have since improved substantially
    at architectural reasoning ($95.4\%$ and $82.4\%$ today,
    Figure~\ref{fig:qa-progress}). The $70\%$ is therefore not competitive now; what
    transfers is that binary Q\&A labels alone lifted a small backbone past the best
    contemporaneous general-purpose model. Whether this specialization also improves
    traffic and buffer-size synthesis remains open.
    
    \section{Additional Performance-Model Results}
    
    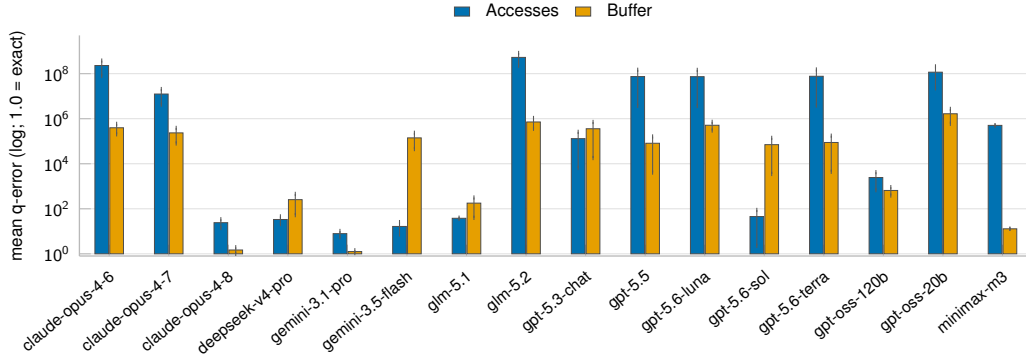
\begin{figure}[ht]
      \centering
\begin{tikzpicture}[font=\sffamily]
\begin{axis}[
  x=0.1313cm, height=4.50cm,
  xmin=-3.00, xmax=93.00,
  ymode=log, log basis y=10,
  ymin=0.8, ymax=5e9,
  ytick={1,1e2,1e4,1e6,1e8},
  yticklabels={$\mathsf{10^{0}}$,$\mathsf{10^{2}}$,$\mathsf{10^{4}}$,$\mathsf{10^{6}}$,$\mathsf{10^{8}}$},
  ylabel={mean q-error (log; 1.0 = exact)},
  ylabel style={font=\sffamily\fontsize{7}{8}\selectfont},
  xtick={0.000,6.000,12.000,18.000,24.000,30.000,36.000,42.000,48.000,54.000,60.000,66.000,72.000,78.000,84.000,90.000},
  xticklabels={{claude-opus-4-6},{claude-opus-4-7},{claude-opus-4-8},{deepseek-v4-pro},{gemini-3.1-pro},{gemini-3.5-flash},{glm-5.1},{glm-5.2},{gpt-5.3-chat},{gpt-5.5},{gpt-5.6-luna},{gpt-5.6-sol},{gpt-5.6-terra},{gpt-oss-120b},{gpt-oss-20b},{minimax-m3}},
  x tick label style={font=\sffamily\fontsize{6.5}{8}\selectfont, rotate=40, anchor=north east},
  y tick label style={font=\sffamily\fontsize{6.5}{8}\selectfont},
  ymajorgrids, grid style={gray!25, line width=0.3pt},
  axis line style={gray!55, line width=0.5pt}, tick style={gray!55},
  axis x line*=bottom, axis y line*=left,
  legend style={at={(0.5,1.02)}, anchor=south, legend columns=6, draw=none,
                fill=none, font=\sffamily\fontsize{7}{8}\selectfont, column sep=0.9ex},
  legend image code/.code={\draw[#1] (0cm,-0.06cm) rectangle (0.20cm,0.06cm);},
]
\addplot[ybar, bar shift=0pt, bar width=5.15pt, fill=okblue, draw=black!70, line width=0.25pt, error bars/.cd, y dir=both, y explicit, error bar style={black!65, line width=0.4pt}, error mark options={mark size=0.8pt, black!65}] coordinates {(-0.750,2.30083e+08) += (0,1.50625e+08) -= (0,1.50625e+08) (5.250,1.24662e+07) += (0,8.25455e+06) -= (0,8.25455e+06) (11.250,24.1752) += (0,10.1595) -= (0,10.1595) (17.250,33.5047) += (0,12.4848) -= (0,12.4848) (23.250,7.87696) += (0,2.4609) -= (0,2.4609) (29.250,16.4528) += (0,9.17172) -= (0,9.17172) (35.250,37.8879) += (0,2.43141) -= (0,2.43141) (41.250,5.27697e+08) += (0,3.01401e+08) -= (0,3.01401e+08) (47.250,130402) += (0,130352) -= (0,123882) (53.250,7.44158e+07) += (0,7.44158e+07) -= (0,7.0695e+07) (59.250,7.35831e+07) += (0,7.3476e+07) -= (0,6.99039e+07) (65.250,45.2313) += (0,43.5864) -= (0,42.9698) (71.250,7.66943e+07) += (0,7.66943e+07) -= (0,7.28596e+07) (77.250,2447.91) += (0,1777.19) -= (0,1777.19) (83.250,1.16437e+08) += (0,9.39413e+07) -= (0,9.39413e+07) (89.250,505484) +- (0,0)};
\addlegendentry{Accesses}
\addplot[ybar, bar shift=0pt, bar width=5.15pt, fill=okorange, draw=black!70, line width=0.25pt, error bars/.cd, y dir=both, y explicit, error bar style={black!65, line width=0.4pt}, error mark options={mark size=0.8pt, black!65}] coordinates {(0.750,396396) += (0,193421) -= (0,193421) (6.750,234901) += (0,155372) -= (0,155372) (12.750,1.48148) += (0,0.481481) -= (0,0.481481) (18.750,254.801) += (0,202.447) -= (0,202.447) (24.750,1.26226) += (0,0.178797) -= (0,0.178797) (30.750,140941) += (0,96034.6) -= (0,96034.6) (36.750,178.486) += (0,138.779) -= (0,138.779) (42.750,717044) += (0,358294) -= (0,358294) (48.750,358773) += (0,358283) -= (0,340834) (54.750,81312.7) += (0,81311.7) -= (0,77247.1) (60.750,512327) += (0,214007) -= (0,214007) (66.750,70472.1) += (0,70470.1) -= (0,66948.5) (72.750,88088.7) += (0,88087.7) -= (0,83684.3) (78.750,651.97) += (0,264.093) -= (0,264.093) (84.750,1.6591e+06) += (0,1.06499e+06) -= (0,1.06499e+06) (90.750,12.9158) +- (0,0)};
\addlegendentry{Buffer}
\end{axis}
\end{tikzpicture}
      \caption{Mean Q-error for memory-access and buffer predictions; 1 is exact
      and lower is better.}
      \label{fig:qerror-4p}
    \end{figure}
    
    \begin{figure}[ht]
      \centering
\begin{tikzpicture}[font=\sffamily]
\begin{axis}[
  x=0.1313cm, height=4.50cm,
  xmin=-3.00, xmax=93.00,
  ymin=0, ymax=110,
  /pgf/number format/assume math mode=true,
  ytick={0,20,40,60,80,100},
  ylabel={exact pass rate (\%)},
  ylabel style={font=\sffamily\fontsize{7}{8}\selectfont},
  xtick={0.000,6.000,12.000,18.000,24.000,30.000,36.000,42.000,48.000,54.000,60.000,66.000,72.000,78.000,84.000,90.000},
  xticklabels={{claude-opus-4-6},{claude-opus-4-7},{claude-opus-4-8},{deepseek-v4-pro},{gemini-3.1-pro},{gemini-3.5-flash},{glm-5.1},{glm-5.2},{gpt-5.3-chat},{gpt-5.5},{gpt-5.6-luna},{gpt-5.6-sol},{gpt-5.6-terra},{gpt-oss-120b},{gpt-oss-20b},{minimax-m3}},
  x tick label style={font=\sffamily\fontsize{6.5}{8}\selectfont, rotate=40, anchor=north east},
  y tick label style={font=\sffamily\fontsize{6.5}{8}\selectfont},
  ymajorgrids, grid style={gray!25, line width=0.3pt},
  axis line style={gray!55, line width=0.5pt}, tick style={gray!55},
  axis x line*=bottom, axis y line*=left,
  legend style={at={(0.5,1.02)}, anchor=south, legend columns=6, draw=none,
                fill=none, font=\sffamily\fontsize{7}{8}\selectfont, column sep=0.9ex},
  legend image code/.code={\draw[#1] (0cm,-0.06cm) rectangle (0.20cm,0.06cm);},
]
\addplot[ybar, bar shift=0pt, bar width=5.15pt, fill=okgreen, draw=black!70, line width=0.25pt, error bars/.cd, y dir=both, y explicit, error bar style={black!65, line width=0.4pt}, error mark options={mark size=0.8pt, black!65}] coordinates {(-1.500,4.11523) +- (0,3.94308) (4.500,16.0494) +- (0,6.37544) (10.500,21.9907) +- (0,11.1032) (16.500,2.21193) +- (0,0.555222) (22.500,19.573) +- (0,7.88636) (28.500,26.1728) +- (0,7.07849) (34.500,3.24074) +- (0,0) (40.500,0) +- (0,0) (46.500,2.16049) +- (0,1.08025) (52.500,13.604) +- (0,9.20994) (58.500,0.0992063) +- (0,0.0992063) (64.500,63.5802) +- (0,9.76931) (70.500,12.3457) +- (0,7.87003) (76.500,3.61111) +- (0,1.01249) (82.500,1.91358) +- (0,1.1303) (88.500,0) +- (0,0)};
\addlegendentry{Both}
\addplot[ybar, bar shift=0pt, bar width=5.15pt, fill=okblue, draw=black!70, line width=0.25pt, error bars/.cd, y dir=both, y explicit, error bar style={black!65, line width=0.4pt}, error mark options={mark size=0.8pt, black!65}] coordinates {(0.000,6.3786) +- (0,4.34351) (6.000,16.0494) +- (0,6.37544) (12.000,21.9907) +- (0,11.1032) (18.000,3.65226) +- (0,0.843292) (24.000,19.573) +- (0,7.88636) (30.000,26.1728) +- (0,7.07849) (36.000,3.24074) +- (0,0) (42.000,0.154321) +- (0,0.154321) (48.000,2.16049) +- (0,1.08025) (54.000,13.604) +- (0,9.20994) (60.000,0.165344) +- (0,0.10417) (66.000,64.6296) +- (0,9.30136) (72.000,12.3457) +- (0,7.87003) (78.000,5.12346) +- (0,1.34973) (84.000,3.33333) +- (0,1.50811) (90.000,0) +- (0,0)};
\addlegendentry{Accesses}
\addplot[ybar, bar shift=0pt, bar width=5.15pt, fill=okorange, draw=black!70, line width=0.25pt, error bars/.cd, y dir=both, y explicit, error bar style={black!65, line width=0.4pt}, error mark options={mark size=0.8pt, black!65}] coordinates {(1.500,58.4877) +- (0,16.4337) (7.500,79.7325) +- (0,13.4057) (13.500,83.9506) +- (0,16.0494) (19.500,63.5288) +- (0,15.2352) (25.500,80.6584) +- (0,8.70943) (31.500,87.8395) +- (0,8.28598) (37.500,71.7593) +- (0,0.925926) (43.500,30.0926) +- (0,21.7593) (49.500,43.5185) +- (0,19.0153) (55.500,92.9843) +- (0,7.01567) (61.500,30.291) +- (0,10.1102) (67.500,87.5) +- (0,8.52094) (73.500,92.3997) +- (0,7.60031) (79.500,52.5309) +- (0,7.67922) (85.500,20.6481) +- (0,8.0586) (91.500,3.7037) +- (0,0)};
\addlegendentry{Buffer}
\end{axis}
\end{tikzpicture}
      \caption{Exact-match decomposition. Buffer capacity is substantially easier
      than memory-access accounting for nearly every model family.}
      \label{fig:breakdown-4p}
    \end{figure}
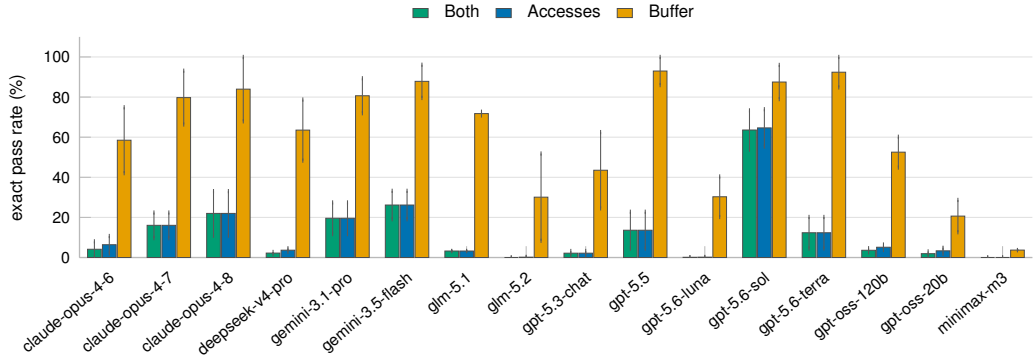
    
    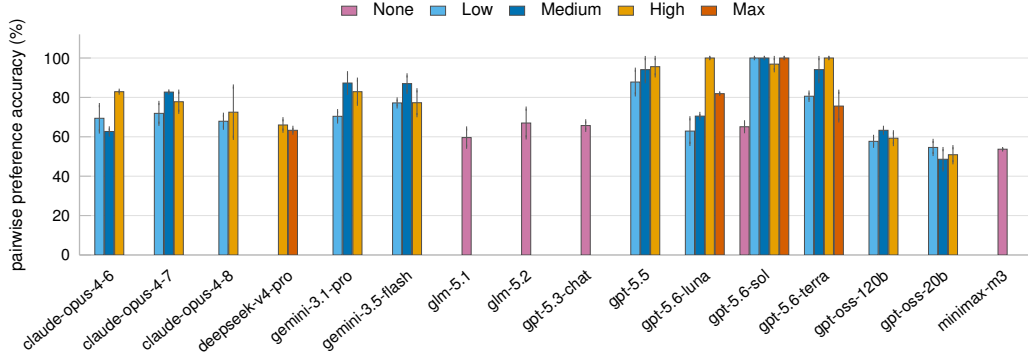
\begin{figure}[ht]
      \centering
\begin{tikzpicture}[font=\sffamily]
\begin{axis}[
  x=0.1313cm, height=4.50cm,
  xmin=-3.00, xmax=93.00,
  ymin=0, ymax=112,
  /pgf/number format/assume math mode=true,
  ytick={0,20,40,60,80,100},
  ylabel={pairwise preference accuracy (\%)},
  ylabel style={font=\sffamily\fontsize{7}{8}\selectfont},
  xtick={0.000,6.000,12.000,18.000,24.000,30.000,36.000,42.000,48.000,54.000,60.000,66.000,72.000,78.000,84.000,90.000},
  xticklabels={{claude-opus-4-6},{claude-opus-4-7},{claude-opus-4-8},{deepseek-v4-pro},{gemini-3.1-pro},{gemini-3.5-flash},{glm-5.1},{glm-5.2},{gpt-5.3-chat},{gpt-5.5},{gpt-5.6-luna},{gpt-5.6-sol},{gpt-5.6-terra},{gpt-oss-120b},{gpt-oss-20b},{minimax-m3}},
  x tick label style={font=\sffamily\fontsize{6.5}{8}\selectfont, rotate=40, anchor=north east},
  y tick label style={font=\sffamily\fontsize{6.5}{8}\selectfont},
  ymajorgrids, grid style={gray!25, line width=0.3pt},
  axis line style={gray!55, line width=0.5pt}, tick style={gray!55},
  axis x line*=bottom, axis y line*=left,
  legend style={at={(0.5,1.02)}, anchor=south, legend columns=6, draw=none,
                fill=none, font=\sffamily\fontsize{7}{8}\selectfont, column sep=0.9ex},
  legend image code/.code={\draw[#1] (0cm,-0.06cm) rectangle (0.20cm,0.06cm);},
]
\addplot[ybar, bar shift=0pt, bar width=3.44pt, fill=okpurple, draw=black!70, line width=0.25pt, error bars/.cd, y dir=both, y explicit, error bar style={black!65, line width=0.4pt}, error mark options={mark size=0.8pt, black!65}] coordinates {(36.000,59.6) +- (0,4.60901) (42.000,67) +- (0,7.29728) (48.000,65.7) +- (0,2.13833) (64.000,65.1) +- (0,2.22565) (90.000,53.7) +- (0,0)};
\addlegendentry{None}
\addplot[ybar, bar shift=0pt, bar width=3.44pt, fill=okcyan, draw=black!70, line width=0.25pt, error bars/.cd, y dir=both, y explicit, error bar style={black!65, line width=0.4pt}, error mark options={mark size=0.8pt, black!65}] coordinates {(-1.000,69.4) +- (0,6.67695) (5.000,71.9) +- (0,5.27408) (11.500,67.9) +- (0,3.26636) (23.000,70.4) +- (0,2.597) (29.000,77.2) +- (0,1.75139) (53.000,87.8) +- (0,6.24075) (58.500,62.9) +- (0,6.49628) (65.000,100) +- (0,0) (70.500,80.6) +- (0,1.85185) (77.000,57.7) +- (0,2.23845) (83.000,54.6) +- (0,3.29843)};
\addlegendentry{Low}
\addplot[ybar, bar shift=0pt, bar width=3.44pt, fill=okblue, draw=black!70, line width=0.25pt, error bars/.cd, y dir=both, y explicit, error bar style={black!65, line width=0.4pt}, error mark options={mark size=0.8pt, black!65}] coordinates {(0.000,62.7) +- (0,1.54321) (6.000,82.7) +- (0,0.308642) (24.000,87.3) +- (0,4.93827) (30.000,87) +- (0,4.17523) (54.000,94.1) +- (0,5.8642) (59.500,70.5) +- (0,1.08025) (66.000,100) +- (0,0) (71.500,94.1) +- (0,5.8642) (78.000,63.3) +- (0,1.23457) (84.000,48.6) +- (0,5.15534)};
\addlegendentry{Medium}
\addplot[ybar, bar shift=0pt, bar width=3.44pt, fill=okorange, draw=black!70, line width=0.25pt, error bars/.cd, y dir=both, y explicit, error bar style={black!65, line width=0.4pt}, error mark options={mark size=0.8pt, black!65}] coordinates {(1.000,82.9) +- (0,0.462963) (7.000,77.8) +- (0,5.0996) (12.500,72.5) +- (0,13.0106) (17.500,66) +- (0,2.89532) (25.000,82.9) +- (0,6.08466) (31.000,77.3) +- (0,6.30605) (55.000,95.6) +- (0,4.39815) (60.500,100) +- (0,0) (67.000,96.9) +- (0,3.08642) (72.500,100) +- (0,0) (79.000,59.3) +- (0,2.97644) (85.000,50.9) +- (0,3.8132)};
\addlegendentry{High}
\addplot[ybar, bar shift=0pt, bar width=3.44pt, fill=okred, draw=black!70, line width=0.25pt, error bars/.cd, y dir=both, y explicit, error bar style={black!65, line width=0.4pt}, error mark options={mark size=0.8pt, black!65}] coordinates {(18.500,63.3) +- (0,1.23457) (61.500,81.9) +- (0,0) (68.000,100) +- (0,0) (73.500,75.6) +- (0,7.25801)};
\addlegendentry{Max}
\end{axis}
\end{tikzpicture}
      \caption{Pairwise preference accuracy induced by generated analytical
      models. Rankings can be correct even when absolute predictions are not.}
      \label{fig:pairwise-4p}
    \end{figure}
    
    \begin{figure}[ht]
      \centering
\begin{tikzpicture}[font=\sffamily]
\begin{axis}[
  width=0.66\linewidth, height=6.33cm,
  xmin=26, xmax=100, ymin=-3, ymax=95,
  /pgf/number format/assume math mode=true,
  xtick={30,40,50,60,70,80,90,100}, ytick={0,20,40,60,80},
  xlabel={overall Q\&A accuracy (\%)},
  ylabel={performance-model exact pass rate (\%)},
  xlabel style={font=\sffamily\fontsize{7}{8}\selectfont},
  ylabel style={font=\sffamily\fontsize{7}{8}\selectfont},
  tick label style={font=\sffamily\fontsize{6.5}{8}\selectfont},
  ymajorgrids, xmajorgrids, grid style={gray!25, line width=0.3pt},
  axis line style={gray!55, line width=0.5pt}, tick style={gray!55},
  axis x line*=bottom, axis y line*=left,
  clip=false,
  legend style={at={(0.5,1.02)}, anchor=south, legend columns=5, draw=none,
                fill=none, font=\sffamily\fontsize{7}{8}\selectfont, column sep=0.9ex},
  legend image code/.code={\draw[#1] (0cm,-0.06cm) rectangle (0.20cm,0.06cm);},
]
\addplot[okred, line width=0.8pt, dashed, forget plot] coordinates {(33.533,-3.000) (96.300,33.679)};
\addplot[only marks, mark=*, mark size=1.7pt, draw=black!55, fill=okpurple, line width=0.3pt] coordinates {(84.2593,4.8000) (29.6296,3.2000) (53.7037,2.2000) (28.7037,0.0000) (29.6296,0.0000)};
\addlegendentry{None}
\addplot[only marks, mark=*, mark size=1.7pt, draw=black!55, fill=okcyan, line width=0.3pt] coordinates {(85.1852,55.2000) (75.0000,8.3000) (82.4074,6.2000) (53.7037,4.6000) (34.2593,2.7000) (30.5556,2.1000) (73.1481,1.2000) (75.9259,1.2000) (46.2963,0.3000) (69.0000,0.2000) (64.8148,0.0000)};
\addlegendentry{Low}
\addplot[only marks, mark=*, mark size=1.7pt, draw=black!55, fill=okblue, line width=0.3pt] coordinates {(87.9630,84.7000) (89.8148,38.4000) (55.5556,17.7000) (85.1852,4.8000) (29.6296,3.4000) (81.9000,0.3000) (31.4815,0.2000) (75.9259,0.0000) (53.7037,0.0000) (81.4815,0.0000)};
\addlegendentry{Medium}
\addplot[only marks, mark=*, mark size=1.7pt, draw=black!55, fill=okorange, line width=0.3pt] coordinates {(85.1852,84.7000) (86.1111,44.2000) (92.5926,40.0000) (87.9630,36.7000) (78.7037,35.6000) (84.2593,29.5000) (68.5185,25.8000) (96.3000,11.9000) (38.8889,4.6000) (31.4815,2.6000) (62.0370,2.2000) (73.1481,0.0000)};
\addlegendentry{High}
\addplot[only marks, mark=*, mark size=1.7pt, draw=black!55, fill=okred, line width=0.3pt] coordinates {(84.2593,88.4000) (78.7037,18.7000) (67.5926,2.2000) (63.8889,0.0000)};
\addlegendentry{Max}
\draw[gray!55, line width=0.3pt] (axis cs:84.2593,88.4000) to[out=0, in=180] (axis cs:102.00,88.40);
\node[anchor=west, font=\sffamily\fontsize{8}{9.5}\selectfont] at (axis cs:103.00,88.40) {gpt-5.6-sol (max)};
\draw[gray!55, line width=0.3pt] (axis cs:87.9630,84.7000) to[out=0, in=180] (axis cs:102.00,82.20);
\node[anchor=west, font=\sffamily\fontsize{8}{9.5}\selectfont] at (axis cs:103.00,82.20) {gpt-5.6-sol (medium)};
\draw[gray!55, line width=0.3pt] (axis cs:85.1852,84.7000) to[out=0, in=180] (axis cs:102.00,76.00);
\node[anchor=west, font=\sffamily\fontsize{8}{9.5}\selectfont] at (axis cs:103.00,76.00) {gpt-5.6-sol (high)};
\draw[gray!55, line width=0.3pt] (axis cs:85.1852,55.2000) to[out=0, in=180] (axis cs:102.00,55.20);
\node[anchor=west, font=\sffamily\fontsize{8}{9.5}\selectfont] at (axis cs:103.00,55.20) {gpt-5.6-sol (low)};
\draw[gray!55, line width=0.3pt] (axis cs:86.1111,44.2000) to[out=0, in=180] (axis cs:102.00,44.20);
\node[anchor=west, font=\sffamily\fontsize{8}{9.5}\selectfont] at (axis cs:103.00,44.20) {gpt-5.5 (high)};
\draw[gray!55, line width=0.3pt] (axis cs:92.5926,40.0000) to[out=0, in=180] (axis cs:102.00,38.00);
\node[anchor=west, font=\sffamily\fontsize{8}{9.5}\selectfont] at (axis cs:103.00,38.00) {gemini-3.5-flash (high)};
\draw[gray!55, line width=0.3pt] (axis cs:89.8148,38.4000) to[out=0, in=180] (axis cs:102.00,31.80);
\node[anchor=west, font=\sffamily\fontsize{8}{9.5}\selectfont] at (axis cs:103.00,31.80) {gemini-3.5-flash (medium)};
\draw[gray!55, line width=0.3pt] (axis cs:87.9630,36.7000) to[out=0, in=180] (axis cs:102.00,25.60);
\node[anchor=west, font=\sffamily\fontsize{8}{9.5}\selectfont] at (axis cs:103.00,25.60) {gemini-3.1-pro (high)};
\draw[gray!55, line width=0.3pt] (axis cs:78.7037,35.6000) to[out=0, in=180] (axis cs:102.00,19.40);
\node[anchor=west, font=\sffamily\fontsize{8}{9.5}\selectfont] at (axis cs:103.00,19.40) {claude-opus-4-8 (high)};
\draw[gray!55, line width=0.3pt] (axis cs:96.3000,11.9000) to[out=0, in=180] (axis cs:102.00,11.90);
\node[anchor=west, font=\sffamily\fontsize{8}{9.5}\selectfont] at (axis cs:103.00,11.90) {claude-opus-4-6 (high)};
\end{axis}
\end{tikzpicture}
      \caption{Q\&A mapping understanding versus analytical-model pass rate across
      configurations ($r=0.46$).}
      \label{fig:correlation-4p}
    \end{figure}
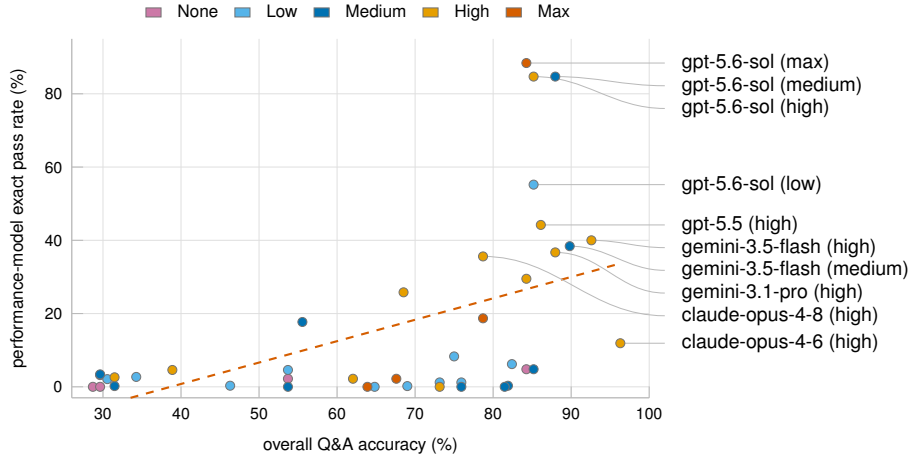
    
    High Q\&A accuracy contains most of the strongest construction results, but the
    spread among high-Q\&A models is large. Conceptual mapping understanding is
    therefore a useful diagnostic, not a sufficient certificate. The exact-match
    decomposition localizes much of the remaining gap to access accounting rather
    than capacity. Pairwise ranking is much stronger than absolute agreement,
    showing that systematic numerical errors can preserve the correct ordering.
    
    \section{Reasoning Effort and Multi-Round Self-Revision}
    
    \begin{figure}[ht]
      \centering
      \input{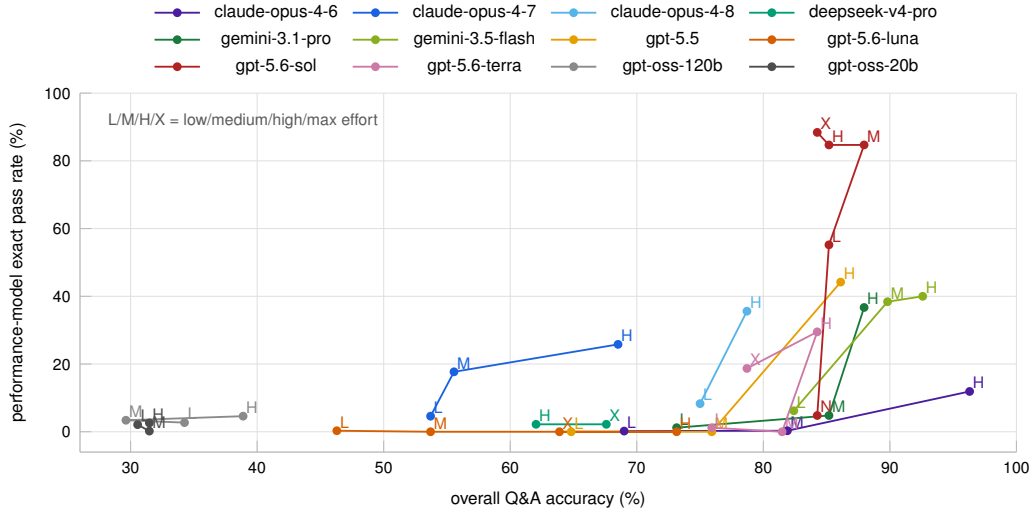}
      \caption{Q\&A and analytical-model accuracy across reasoning-effort settings.
      Additional effort is not uniformly beneficial.}
      \label{fig:effort-4p}
    \end{figure}
    
    \begin{figure}[ht]
      \centering
\begin{tikzpicture}[font=\sffamily]
\begin{axis}[
  name=panel0,
  scale only axis=true,
  width=0.38\linewidth, height=2.30cm,
  xmin=-0.25, xmax=3.25, ymin=-4, ymax=100,
  xtick={0,1,2,3}, ytick={0,20,40,60,80,100},
  /pgf/number format/assume math mode=true,
  xlabel={self-revision round $k$},
  xlabel style={font=\sffamily\fontsize{7.5}{9}\selectfont},
  ylabel style={font=\sffamily\fontsize{7.5}{9}\selectfont},
  tick label style={font=\sffamily\fontsize{7}{8.5}\selectfont},
  title={(a) \textsf{gpt-5.6-sol} --- strongest closed},
  title style={font=\sffamily\fontsize{7.5}{9}\selectfont, yshift=-2pt},
  ymajorgrids, grid style={gray!25, line width=0.3pt},
  axis line style={gray!55, line width=0.5pt}, tick style={gray!55},
  axis x line*=bottom, axis y line*=left,
  ylabel={both-exact pass rate (\%)},
  legend style={at={(1.10,1.20)}, anchor=south, legend columns=4,
                draw=none, fill=none, font=\sffamily\fontsize{7.5}{9}\selectfont, column sep=0.9ex},
  legend image code/.code={\draw[#1, solid] (0cm,-0.06cm) rectangle (0.20cm,0.06cm);},
]
\addplot[okcyan, line width=0.9pt, mark=*, mark size=1.5pt, mark options={solid, fill=okcyan, draw=okcyan}] coordinates {(0,88.4259) (1,88.4259) (2,0.0000) (3,0.0000)};
\addlegendentry{Low}
\addplot[okblue, line width=0.9pt, mark=*, mark size=1.5pt, mark options={solid, fill=okblue, draw=okblue}, dash pattern=on 4pt off 1.6pt] coordinates {(0,88.4259) (1,88.4259) (2,88.4259) (3,88.4259)};
\addlegendentry{Medium}
\addplot[okorange, line width=0.9pt, mark=*, mark size=1.5pt, mark options={solid, fill=okorange, draw=okorange}, dash pattern=on 2.2pt off 1.8pt] coordinates {(0,88.4259) (1,88.4259) (2,88.4259) (3,88.4259)};
\addlegendentry{High}
\addplot[okred, line width=0.9pt, mark=*, mark size=1.5pt, mark options={solid, fill=okred, draw=okred}, dash pattern=on 1.2pt off 1.6pt] coordinates {(0,0.0000) (1,0.0000) (2,0.0000) (3,88.4259)};
\addlegendentry{Max}
\end{axis}
\begin{axis}[
  name=panel1,
  at={(panel0.east)}, anchor=west, xshift=1.0cm,
  scale only axis=true,
  width=0.38\linewidth, height=2.30cm,
  xmin=-0.25, xmax=3.25, ymin=-4, ymax=100,
  xtick={0,1,2,3}, ytick={0,20,40,60,80,100},
  /pgf/number format/assume math mode=true,
  xlabel={self-revision round $k$},
  xlabel style={font=\sffamily\fontsize{7.5}{9}\selectfont},
  ylabel style={font=\sffamily\fontsize{7.5}{9}\selectfont},
  tick label style={font=\sffamily\fontsize{7}{8.5}\selectfont},
  title={(b) \textsf{gpt-oss-20b} --- strongest open-weight},
  title style={font=\sffamily\fontsize{7.5}{9}\selectfont, yshift=-2pt},
  ymajorgrids, grid style={gray!25, line width=0.3pt},
  axis line style={gray!55, line width=0.5pt}, tick style={gray!55},
  axis x line*=bottom, axis y line*=left,
  yticklabels={},
]
\addplot[okcyan, line width=0.9pt, mark=*, mark size=1.5pt, mark options={solid, fill=okcyan, draw=okcyan}, forget plot] coordinates {(0,0.0000) (1,0.0000) (2,0.0000) (3,0.0000)};
\addplot[okblue, line width=0.9pt, mark=*, mark size=1.5pt, mark options={solid, fill=okblue, draw=okblue}, dash pattern=on 4pt off 1.6pt, forget plot] coordinates {(0,0.0000) (1,0.0000) (2,15.7407) (3,0.0000)};
\addplot[okorange, line width=0.9pt, mark=*, mark size=1.5pt, mark options={solid, fill=okorange, draw=okorange}, dash pattern=on 2.2pt off 1.8pt, forget plot] coordinates {(0,12.9630) (1,3.2407) (2,3.2407) (3,3.2407)};
\end{axis}
\end{tikzpicture}
      \caption{Feedback-free self-revision may help, harm, or leave accuracy
      unchanged for the strongest closed and open-weight models.}
      \label{fig:revision-4p}
    \end{figure}
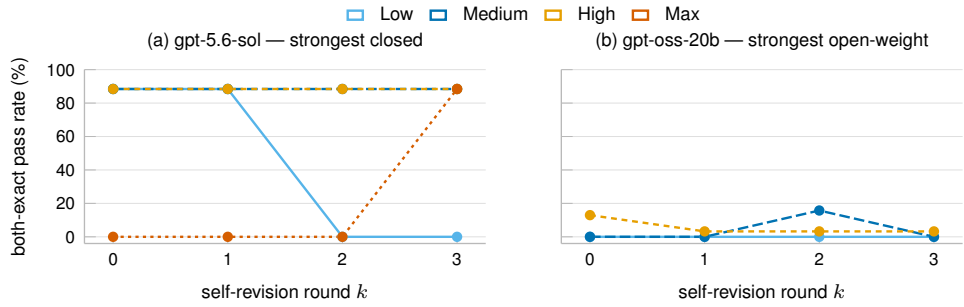
    
    Each model may revise its initial program up to three times without execution
    or pass/fail feedback. Across 41 configurations, 18 remain unchanged through
    all rounds, 10 finish below their initial accuracy, and only 7 finish above it.
    Mean pass rate rises only from $14.0\%$ to $16.7\%$, although selecting the best
    candidate in hindsight would reach $22.9\%$. Models can therefore generate a
    better candidate without reliably recognizing and retaining it.
    
    \section{Per-Round Results for Multi-Round Self-Revision}
    \label{app:iterative}
    
    Section~\ref{sec:results} summarizes the effect of feedback-free multi-round
    self-revision on
    analytical performance modeling. Table~\ref{tab:kiter_allrounds} gives the full
    per-round breakdown behind that summary: for every (model, reasoning-effort)
    configuration, the both-exact pass rate of the program produced at each revision
    step $k$, together with the best rate achieved over all four candidates.
    
    \begin{table}[ht!]
    \centering
    \caption{%
      \textbf{Both-exact pass rate (\%) across three self-revision rounds on performance modeling.}
      Each row denotes one (model, reasoning-effort) configuration; the model generates a program at $k{=}0$ and is given the option to revise up to three times with no evaluation feedback.
    }
    \label{tab:kiter_allrounds}
    \scriptsize
    \begin{tabular}{@{}ll rrrr r@{}}
    \toprule
    Model & Effort & $k{=}0$ & $k{=}1$ & $k{=}2$ & $k{=}3$ & Best \\
    \midrule
    \multirow{3}{*}{opus-4.6}
      & Low     &  0.0 &  0.0 &  0.0 &  0.0 & \textbf{ 0.0} \\
      & Medium  &  7.9 &  7.9 & 35.6 & 35.6 & \textbf{35.6} \\
      & High    &  0.0 &  0.0 &  0.0 &  0.0 & \textbf{ 0.0} \\[2pt]
    \multirow{3}{*}{opus-4.7}
      & Low     & 11.1 & 13.9 & 13.9 & 13.9 & \textbf{13.9} \\
      & Medium  & 41.7 & 45.4 & 41.7 & 41.7 & \textbf{45.4} \\
      & High    & 25.9 & 35.6 & 35.6 & 35.6 & \textbf{35.6} \\[2pt]
    \multirow{2}{*}{opus-4.8}
      & Low     & 18.5 & 18.5 & 18.5 & 18.5 & \textbf{18.5} \\
      & High    & 29.6 & 43.5 & 43.5 & 43.5 & \textbf{43.5} \\[2pt]
    \multirow{2}{*}{deepseek-v4-pro}
      & High    &  3.2 &  0.0 &  0.0 &  0.0 & \textbf{ 3.2} \\
      & Max     &  3.2 &  3.2 &  0.0 &  3.2 & \textbf{ 3.2} \\[2pt]
    \multirow{3}{*}{gemini-3.1-pro}
      & Low     &  3.7 &  3.2 &  3.2 &  3.2 & \textbf{ 3.7} \\
      & Medium  & 80.1 & 80.1 & 80.1 & 80.1 & \textbf{80.1} \\
      & High & 22.5 & 22.5 & 22.5 &  1.2 & \textbf{22.5} \\[2pt]
    \multirow{3}{*}{gemini-3.5-flash}
      & Low     &  7.4 &  7.4 &  7.4 &  0.0 & \textbf{ 7.4} \\
      & Medium  & 23.6 & 23.6 & 23.6 & 23.6 & \textbf{23.6} \\
      & High    &  3.7 &  3.7 &  3.7 & 35.6 & \textbf{35.6} \\[2pt]
    glm-5.1      & Default &  0.0 &  0.0 &  0.0 &  0.0 & \textbf{ 0.0} \\[2pt]
    glm-5.2      & Default &  0.0 &  0.0 &  0.0 &  0.0 & \textbf{ 0.0} \\[2pt]
    gpt-5.3      & Default &  3.2 &  0.0 &  0.0 &  1.4 & \textbf{ 3.2} \\[2pt]
    \multirow{3}{*}{gpt-5.5}
      & Low     &  0.0 & 77.3 & 77.3 & 77.3 & \textbf{77.3} \\
      & Medium  &  0.0 &  0.0 &  0.0 &  0.0 & \textbf{ 0.0} \\
      & High    &  0.0 & 88.4 &  0.0 &  0.0 & \textbf{88.4} \\[2pt]
    \multirow{4}{*}{gpt-5.6-luna}
      & Low     &  1.4 &  0.0 &  0.0 &  0.0 & \textbf{ 1.4} \\
      & Medium  &  0.0 &  0.0 &  0.9 &  0.0 & \textbf{ 0.9} \\
      & High    &  0.0 &  0.0 &  0.0 &  0.0 & \textbf{ 0.0} \\
      & Max     &  0.0 &  0.0 &  0.0 &  0.0 & \textbf{ 0.0} \\[2pt]
    \multirow{5}{*}{gpt-5.6-sol}
      & Low     & 88.4 & 88.4 &  0.0 &  0.0 & \textbf{88.4} \\
      & Medium  & 88.4 & 88.4 & 88.4 & 88.4 & \textbf{88.4} \\
      & High    & 88.4 & 88.4 & 88.4 & 88.4 & \textbf{88.4} \\
      & Max     &  0.0 &  0.0 &  0.0 & 88.4 & \textbf{88.4} \\[2pt]
    \multirow{4}{*}{gpt-5.6-terra}
      & Low     &  0.0 &  0.0 &  0.0 &  0.0 & \textbf{ 0.0} \\
      & Medium  &  3.7 &  0.0 &  0.0 &  0.0 & \textbf{ 3.7} \\
      & High    &  0.0 &  0.0 &  3.7 &  0.0 & \textbf{ 3.7} \\
      & Max     &  3.7 &  0.0 &  0.0 &  0.0 & \textbf{ 3.7} \\[2pt]
    \multirow{3}{*}{gpt-oss-20b}
      & Low     &  0.0 &  0.0 &  0.0 &  0.0 & \textbf{ 0.0} \\
      & Medium  &  0.0 &  0.0 & 15.7 &  0.0 & \textbf{15.7} \\
      & High    & 13.0 &  3.2 &  3.2 &  3.2 & \textbf{13.0} \\[2pt]
    \multirow{3}{*}{gpt-oss-120b}
      & Low     &  3.2 &  3.2 &  3.2 &  3.2 & \textbf{ 3.2} \\
      & Medium  &  0.0 &  0.0 &  0.0 &  0.0 & \textbf{ 0.0} \\
      & High    &  0.0 &  0.0 &  0.0 &  0.0 & \textbf{ 0.0} \\[2pt]
    minimax-m3   & Default &  0.0 &  0.0 &  0.0 &  0.0 & \textbf{ 0.0} \\
    \bottomrule
    \end{tabular}
    \end{table}
    
    \section{Limitations and Extensions}
    
    The benchmark measures agreement with Timeloop and its accounting conventions,
    not physical accuracy on silicon. It fixes one two-level hierarchy and omits
    spatial tiling, deeper hierarchies, sparse and attention workloads, energy,
    latency, and area. Model versions changed during the evaluation window,
    providers expose reasoning effort differently, and replicate counts vary.
    
    A natural extension is to evaluate measured GPUs, where DRAM traffic,
    occupancy, and kernel time replace tool-derived labels. Such an evaluation
    would require tolerance bands calibrated to measurement noise and controlled
    pairs that isolate coalescing, cache behavior, occupancy, and latency hiding.
    It would also test whether the controlled mapping-reasoning findings transfer
    from analytical ground truth to shipping hardware.
    from analytical reference labels to shipping hardware.
    
    \end{document}